\RequirePackage{fix-cm} 
\documentclass{svjour3}                     
\smartqed  
\usepackage{amsmath,verbatim,amssymb,amsfonts,amscd, graphicx,booktabs,multirow,graphics,xcolor,hyperref}
\usepackage{listings}
\usepackage{mathtools} 

\def\clqq{``}
\def\crqq{''}
\def\quo#1{\clqq{}#1\crqq{}}  

\usepackage{comment}
\usepackage{url}
\usepackage{booktabs}
\usepackage{algorithmic}
\usepackage{algorithm}
\usepackage{comment}
\usepackage{mwe} 
\usepackage{longtable,booktabs}

\usepackage{amsthm} 
\begin{document}
\title{Multi-task twin support vector machine with Universum data}

\titlerunning{Multi-task Twin support vector machine}        

\author{Hossein Moosaei$^*$ \and
	Fatemeh Bazikar	\and
	 Milan Hlad\'{i}k}


\institute{ $*$ Corresponding Author \at
	 	\and
	Hossein Moosaei \at
	Department of Mathematics, University of Bojnord, Bojnord, Iran\\
		Department of Applied Mathematics, School of Computer Science, Charles University, Prague, Czech Republic \\
	\email{hmoosaei@gmail.com, hmoosaei@kam.mff.cuni.cz}
	\and
	Fatemeh Bazikar\at
	Department of Applied Mathematics, Faculty of Mathematical Sciences,  University of Guilan, Rasht, Iran \\
	\email{f.bazikar@gmail.com, fatemeh$_{-}$bazikar@phd.guilan.ac.ir}
		\and 
		 Milan Hlad\'{i}k  \at
	Department of Applied Mathematics, Faculty  of  Mathematics  and  Physics, Charles University, Prague, Czech Republic \\
	\email{hladik@kam.mff.cuni.cz}  
}

\date{Received: date / Accepted: date}
\maketitle
\begin{abstract}
Multi-task learning (MTL) has emerged as a promising topic of machine learning in recent years, aiming to enhance the performance of numerous related learning tasks by exploiting beneficial information. During the training phase, most of the existing multi-task learning models concentrate entirely on the target task data and ignore the non-target task data contained in the target tasks. To address this issue, Universum data, that do not correspond to any class of a classification problem, may be used as prior knowledge in the training model. This study looks at the challenge of multi-task learning using Universum data to employ non-target task data, which leads to better performance. It proposes a multi-task twin support vector machine with Universum data ($\mathfrak{U}$MTSVM) and provides two approaches to its solution. The first approach takes into account the dual formulation of $\mathfrak{U}$MTSVM and tries to solve a quadratic programming problem. The second approach formulates a least-squares version of $\mathfrak{U}$MTSVM and refers to it as LS-$\mathfrak{U}$MTSVM to further increase the generalization performance. The solution of the two primal problems in LS-$\mathfrak{U}$MTSVM is simplified to solving just two systems of linear equations, resulting in an incredibly simple and quick approach.
Numerical experiments on several popular multi-task data sets and medical data sets demonstrate the efficiency of the proposed methods.
\end{abstract}

\section{Introduction}
Pattern recognition is becoming more important in various fields due to the development of machine learning. Traditional pattern recognition focuses on single-task learning (STL), with multi-task learning (MTL) generally being disregarded. The MTL aims to use helpful information in several related tasks to improve the generalization performance of all tasks. Multi-task learning aims to enhance predictions by exchanging group knowledge amongst related training data sets known as ``tasks''.
Therefore, multi-task learning is a significant area of research in machine learning. 
The study of multi-task learning has been of interest in diverse fields, including multi-level analysis~\cite{bakker2003task}, medical diagnosis~\cite{bi2008improved}, semi-supervised learning~\cite{ando2005framework},  web search ranking~\cite{chapelle2010multi}, speech recognition~\cite{birlutiu2010multi}, cell biology~\cite{ren2016multicell}, person identification~\cite{su2017multi}, drug interaction extraction~\cite{zhou2018position},  object tracking~\cite{cheng2015multi}, etc.

Several MTL approaches have been proposed throughout the years, which can be classified into several groups. SVM-based methods are among such approaches. Because of the effectiveness of support vector machine (SVM) \cite{burges1998tutorial} in multi-task learning, several researchers have concentrated on multi-task SVM \cite{ji2013multitask,shiao2012implementation,xue2016multi,yang2010multi}. Evgeniou et al.~\cite{evgeniou2004regularized} developed a multi-task learning strategy based on the minimization of regularization functions similar to those used in SVM.

As we know, Jeyadeva et al.~\cite{khemchandani2007twin} proposed twin support vector machine (TSVM) for binary classification in 2007, based on the main notion of GEPSVM \cite{mangasarian2005multisurface}. TSVM divides the positive and negative samples by producing two non-parallel hyperplanes via solving two smaller quadratic programming problems (QPP) rather than one large QPP considered in SVM. In contrast to the substantial research conducted on multi-task support vector machines, there have been few efforts to incorporate multi-task learning into twin support vector machines (TSVM). For instance, inspired by multi-task learning and TSVM, Xie and Sun proposed a multi-task twin support vector machine \cite{xie2012multitask}. They used twin support vector machines for multi-task learning and referred to the resulting model as directed multi-task twin support vector machine (DMTSVM). Following that, multi-task centroid twin support vector machines (MCTSVM) \cite{xie2015multitask} were suggested to cope with outlier samples in each task. In addition, motivated by least-squares twin support vector machine (LS-TWSVM) \cite{kumar2009least}, Mei and Xu~\cite{mei2019multi} proposed a multi-task least-squares twin support vector machine (MTLS-TWSVM). Instead of the two QPP problems addressed by DMTSVM, MTLS-TWSVM solves only two smaller linear equations, resulting in quick computation. 

Universum data is defined as a set of unlabeled samples that do not belong to any class \cite{chapelle2007analysis,vapnik200624,weston2006inference}. These data demonstrate the ability to encode past knowledge by providing meaningful information in the same domain as the problem. The Universum data have effectively improved learning performance in classification and clustering.  By incorporating the Universum data into SVM, Vapnik \cite{vapnik2006estimation} proposed a novel model and referred to it as support vector machine with Universum ($\mathfrak{U}$-SVM). Weston et al.~\cite{weston2006inference} investigated this new framework (Universum data) and proved that the use of these data outperformed approaches that just used labeled samples. 

Inspired by $\mathfrak{U}$-SVM, Sinz et al.~\cite{chapelle2007analysis} introduced  least-squares support vector machine with Universum data ($\mathfrak{U}_{LS}$-SVM). Also, Zhang et al.~\cite{zhang2008semi} proposed semi-supervised algorithms based on a graph for the learning of labeled samples, unlabeled samples, and Universum data. Various studies confirm the helpfulness of Universum data for supervised and semi-supervised learning. The training procedure incorporates Universum data, which increases the total number of samples and adds substantial computing complexity. As a result, the classical $\mathfrak{U}$-SVM has disadvantages such as high computational complexity due to facing a larger QPP.
Fortunately, in 2012, Qi et al.~\cite{qi2012twin} proposed a twin support vector machine using Universum data ($\mathfrak {U}$-TSVM) that addressed this computational shortcoming. Instead of solving one large QPP, which is done in the standard $\mathfrak {U}$-SVM algorithm, this approach solves two smaller QPPs. The authors demonstrated that the approach not only reduced the time of computation, but also outperformed the traditional $\mathfrak{U}$-SVM in terms of classification accuracy.  Following that, Xu et al.~\cite{xu2016least} developed  least-squares twin support vector machine using Universum data ($\mathfrak{U}_{LS}$-TSVM) for classification. They described the way two nonparallel hyperplanes could be found by solving a pair of systems of linear equations. As a result, $\mathfrak{U}_{LS}$-TSVM works faster than $\mathfrak{U}$-TSVM. Inspired by $ \nu$-TSVM and $\mathfrak{U}$-TSVM, Xu et al.~\cite{xu2016nu} presented a $ \nu$-TSVM  with Universum data ($\mathfrak{U}_{\nu}$-TSVM).  It allows the incorporation of the prior knowledge embedded in the unlabeled samples into supervised learning to improve generalization performance. Xiao et al.~\cite{xiao2021new} established Universum learning in 2021 to make non-target task data behave as previous knowledge, and suggested a novel multi-task support vector machine using Universum data (U-MTLSVM). In general, Universum models have been of interest to many researchers because of their simple structures and good generalization performance \cite{moosaei2021DC,moosaei2021sparse,richhariya2018improved,richhariya2018eeg}.

Despite the work done in MTL, there is still a need to create more efficient approaches in terms of accuracy and other field measures. Inspired by DMTSVM and Universum data, we present a significant multi-task twin support vector machine using Universum data ($ \mathfrak{U} $MTSVM).
This paper presents two approaches to the solution of the proposed model.
We obtain the dual formulation of $\mathfrak{U}$MTSVM, and try to solve the quadratic programming problems at the first approach. In addition, we propose a least-squares version of the multi-task twin support vector machine with Universum data (referred to as LS-$\mathfrak{U}$MTSVM) to further increase the generalization performance and reduce the time of computation. The LS-$\mathfrak{U}$MTSVM only deals with two smaller linear equations instead of the two dual quadratic programming problems which are used in $ \mathfrak{U} $MTSVM. 

The contributions of our research can be summarized as follows. 
\begin{itemize}
	\item  Using Universum data, we present a new multi-task twin support vector machine model. This model naturally extends DMTSVM by adding Universum data.
	\item The proposed model has the same advantages as DMTSVM, and furthermore, improves its performance.
	\item We propose two approaches to finding the solution of the proposed model, namely, solving the dual problem instead of the
	primal problem, and introducing the least-squares version of $\mathfrak{U}$MTSVM, which is called LS-$\mathfrak{U}$MTSVM, for solving the primal problem.
\end{itemize}

The remainder of this paper is organized as follows. Following a quick review of
DMTSVM and MTLS-TWSVM in Section~2, we describe the details of our proposed $\mathfrak{U}$MTSVM and introduce its dual problem in Section~3. In Section~4, we present a new algorithm, namely, LS-$\mathfrak{U}$MTSVM. Next, we provide some numerical experiments in Section~5. Finally, we summarize our findings in Section~6 within a brief conclusion.

\paragraph{Notation.}
We use $ \mathbb{R}^n $ for  the $ n$-dimensional real vector space and $I$ for the identity matrix. The transpose and Euclidean norm of a matrix $ A $ are denoted by the symbols $A^T $ and $ \|\cdot\| $, respectively. 
    The gradient of the function $f$ with respect to the variable $ x $ is denoted by $ \nabla_{x} f(x)$ or simply $ \nabla f(x)$.
  Next, we use $\langle x,y\rangle =x^{T}y  $ to denote the inner product of two  $ n$-dimensional vectors $ x $ and $ y $. The symbol $blkdiag({{P}_{1}},\ldots,{{P}_{T}})$ denotes the block-diagonal matrix created by ${{P}_{1}},\ldots,{{P}_{T}}$ matrices.

\section{Related work}
In this section, we first give an overview of the multi-task problem. Then, we introduce the direct multi-task twin support vector machine (DMTSVM)  and the multi-task least-squares twin support vector machine (MTLS-TWSVM). It is preferable to define the fundamental foundations of these procedures since they serve as a solid basis for our suggested method.  For the multi-task problem, we have  $T$ training task, and we assume the set $S_{t}$, for $t=1,\ldots, T$, stores  the labeled  samples for the $t$-th task. It is given by 
$$S_{t}=\{(x_{1t}, y_{1t}),\ldots,(x_{{n_{t}}t}, y_{{n_{t}}t}) \}
,$$
where, $n_{t}$ is the number of samples in task $t$, $x_{it}\in \mathbb{R}^{n}$, $y_{it}\in \{\pm 1\},\,\, i=1,\ldots,n_{t}$.

\subsection{Multi-task twin support vector machine}\label{2.1} 

Xie et al.~\cite{xie2012multitask} introduced a new classification method that directly incorporates the regularized multi-task learning (RMTL) \cite{evgeniou2004regularized}  concept into TSVM and called it direct multi-task twin support vector machine (DMTSVM). 

Suppose positive samples and positive samples in 
$t$-th task are presented by $X_{p}$ and $X_{pt}$, respectively, while $X_{n}$  represents the negative samples and negative samples in $t$-th task are presented by $X_{nt}$,  
 that is, $X_p^T=[X_{p1}\,X_{p2}\, \dots X_{pT}]$.
Now, for every task $t\in \{1, \ldots, T\}$ we define:
\[ A=[X_{p}\,\,\, e_1],~ A_{t}=[X_{pt}\,\,\, e_{1t}],~ B=[X_{n}\,\,\, e_2],~ B_{t}=[X_{nt}\,\,\, e_{2t}], \]
where $e_1,e_2,e_{1t}$ and $e_{2t}$ are one vectors of appropriate dimensions. Assume that all tasks have two mean hyperplanes in common, i.e., $u_{0}=[w_{1}, b_{1}]^{T}$ and $v_{0}=[w_{2}, b_{2}]^{T}$. The two hyperplanes in the $t$-th task for positive and negative classes are 
$(u_{0}+u_{t})=[w_{1t}, b_{1t}]^{T}$  and $(v_{0}+v_{t})=[w_{2t}, b_{2t}]^{T}$, respectively. The bias between task $t$  and the common mean vectors  $u_{0}$ and $v_{0}$ is represented by $u_{t}$ and $v_{t}$, respectively.

The DMTSVM optimization problems are expressed below:
\begin{align}\label{1} 
\underset{u_{0}, u_{t}, \xi_{t}}\min\ 
 &\dfrac{1}{2}\| Au_{0}\|^{2}
  +\dfrac{\mu_{1}}{2T}\sum_{t=1}^{T} \| A_{t}u_{t}\|^{2}+c_{1}\sum_{t=1}^{T}e_{2t}^{T}\xi_{t},\nonumber\\
\text{s.t.}\,\,\,\,& -B_{t}(u_{0}+u_{t})+\xi_{t}\geq e_{2t},\nonumber\\
&\hspace*{2.7cm} \xi_{t}\geq 0,
\end{align}
and
\begin{align}\label{2} 
\underset{v_{0}, v_{t}, \eta_{t}}\min\ &\dfrac{1}{2}\| Bv_{0}\|^{2}+\dfrac{\mu_{2}}{2T}\sum_{t=1}^{T} \| B_{t}v_{t}\|^{2}+c_{2}\sum_{t=1}^{T}e_{1t}^{T}\eta_{t},\nonumber\\
\text{s.t.}\,\,\,\,& A_{t}(v_{0}+v_{t})+\eta_{t}\geq e_{1t},\nonumber\\
&\hspace*{2.2cm} \eta_{t}\geq 0.
\end{align}
 In problems \eqref{1} and \eqref{2},  $t\in\{1,\ldots,T\}$, $c_{1}$, $c_{2}$, and $e_{1t}$ and $e_{2t}$ are one vectors of appropriate dimensions.  Next, $\mu_{1}$ and $\mu_{2}$  are positive parameters used for correlation of all tasks.
 If  $\mu_{1}$ and $\mu_{2}$ give small penalty on vectors $ u_{t} $ and $ v_{t} $, then $ u_{t} $ and $ v_{t} $ tend to be larger. As a consequence, the models give less similarity.
 When $\mu_{1}\rightarrow \infty$ and $\mu_{2}\rightarrow \infty$, $ u_{t} $ and $ v_{t} $ tend to be smaller and make the $T$ models similar~\cite{evgeniou2004regularized}.

By defining
\begin{align*}
& Q =B{{({{A}^{T}}A)}^{-1}}{{B}^{T}}, \quad  
 {{P}_{t}}={{B}_{t}}{{(A_{t}^{T}A_{t})}^{-1}}B_{t}^{T}, \\
& \alpha=[\alpha_{1}^{{T}},\ldots,\alpha_{T}^{{T}}]^{T},\quad
 P=blkdiag({{P}_{1}},\ldots,{{P}_{T}}),  
\end{align*}
the dual problem of the problem \eqref{1} may be expressed as follows:
\begin{align}\label{13} 
\underset{\alpha}\max&\  -\dfrac{1}{2}\alpha^{T}(Q+\textstyle\frac{T}{\mu_{1}}P)\alpha
 +e_{2}^T\alpha\nonumber\\
\text{s.t.}\,& \ \ 0\leq\alpha\leq c_{1}e_{2}.
\end{align}
By resolving the aforementioned dual problem, we may discover:
\begin{align} \nonumber
& u_{0}= -(A^{T}A)^{-1}B^{T}\alpha, \\  
& u_{t}= -\dfrac{T}{\mu_{1}}(A^{T}_{t}A_{t})^{-1}B^{T}_{t}\alpha_{t}. \nonumber
\end{align}

Similarly, we may derive the dual problem of the problem \eqref{2} as follows:
\begin{align}\label{14} 
\underset{\alpha^{\ast}}\max&\  -\dfrac{1}{2}\alpha^{\ast^{T}}(R+\textstyle\frac{T}{\mu_{2}}S)\alpha^{\ast}
+e_{1}^T\alpha^{\ast}\nonumber\\
\text{s.t.}\,& \ \ 0\leq\alpha^{\ast}\leq c_{2}e_{1},
\end{align}
where $\alpha^{\ast}=[\alpha_{1}^{\ast^{T}},\ldots,\alpha_{T}^{\ast^{T}}]^{T}$,
$R=A{{({{B}^{T}}B)}^{-1}}{{A}^{T}}$, and ${{S}_{t}}={{A}_{t}}{{(B_{t}^{T}{{B}_{t}})}^{-1}}A_{t}^{T}$ and $S=blkdiag({{S}_{1}},\ldots ,{{S}_{T}})$.
By solving problem \eqref{13} and \eqref{14}, we can set the hyperplanes of every task $(u_{0}+u_{t})$ and 
$(v_{0}+v_{t})$. Meanwhile, a new data point $x$ in the $t$-th  task is determined to class $i\in \{+1, -1\}$ by using the following decision function:
\begin{align}\label{15} 
f(x)=\arg \underset{k=1,2}\min\, |x^{T}w_{kt}+b_{kt}|.
\end{align}

\subsection{Multi-task least squares twin support vector machine}\label{2.2} 
Inspired by DMTSVM and the least squares twin support vector machine (LSTWSVM), Mei et al.~\cite{mei2019multi} proposed  a novel multi-task least squares twin support vector machine and called MTLS-TWSVM. 

The notation of $X_{p}, X_{pt},X_{n},X_{nt}, A, A_{t}, B$ and $B_{t}$ is the same as that used in subsection \eqref{2.1}. The MTLS-TWSVM problems are formulated as follows:
\begin{align}\label{16} 
\mathop {\min }_{u_{0}, u_{t}, \xi_{t}\,}  \, &\dfrac{1}{2}\| Au_{0}\|^{2} +\dfrac{\mu_{1}}{2T}\sum_{t=1}^{T} \| A_{t}u_{t}\|^{2}+\dfrac{c_{1}}{2}\sum_{t=1}^{T}\| \xi_{t}\|^{2},\nonumber\\
\text{ s.t.} ~~& -B_{t}(u_{0}+u_{t})+\xi_{t}=e_{2t},
\end{align}
and
\begin{align}\label{17} 
\mathop {\min }_{v_{0}, v_{t}, \eta_{t}\,}  \,&   \dfrac{1}{2}\| Bv_{0}\|^{2} +\dfrac{\mu_{2}}{2T}\sum_{t=1}^{T} \| B_{t}v_{t}\|^{2}+\dfrac{c_{2}}{2}\sum_{t=1}^{T}\|\eta_{t}\|^{2},\nonumber\\
\text{ s.t.} ~~&  A_{t}(v_{0}+v_{t})+\eta_{t}= e_{1t},
\end{align}
where $\mu_{1}, c_{1},\mu_{2}$ and $c_{2}$ are positive parameters. 
The Lagrangian function associated with the problem \eqref{16} is defined by
\begin{align}\label{18} 
L_{1}= \dfrac{1}{2}\| Au_{0}\|^{2}&+\dfrac{\mu_{1}}{2T}\sum_{t=1}^{T} \parallel A_{t}u_{t}\parallel^{2}+\dfrac{c_{1}}{2}\sum_{t=1}^{T}\| \xi_{t}\parallel^{2}\nonumber\\
& -\sum_{t=1}^{T}\alpha_{t}^{T}\big(- B_{t}(u_{0}+u_{t})+\xi_{t}-e_{2t}\big),
\end{align}
where  $\alpha=[\alpha_{1}^{T},\ldots,\alpha_{T}^{T}]^{T}$ are the Lagrangian multipliers. 
After writing the partial derivatives of Lagrangian function \eqref{18} with respect to $ u_{0}$, $ u_{t} $, $ \xi_{t} $, and $ \alpha_{t} $,  we derive 
\begin{equation}\nonumber
\alpha = \left(Q+\frac{T}{\mu_{1}}P+\frac{1}{c_{1}}I\right)^{-1}e_{2},
\end{equation}
where $Q=B(A^{T}A)^{-1}B^{T}$, $P_{t}=B_{t}(A_{t}^{T}A_{t})^{-1}B^{T}_{t}$, and
$P=blkdiag(P_{1},\ldots,P_{T})$. 
Then we can compute  the solution of problem~\eqref{16}:
\begin{align} \nonumber
u_{0}=-(A^{T}A)^{-1}B^{T}\alpha,~~~
 u_{t}=-\dfrac{T}{\mu_{1}}(A^{T}_{t}A_{t})^{-1}B^{T}_{t}\alpha_{t}.
\end{align}


Similarly, the following relations may be used to find the solution of \eqref{17}:
\begin{align}\nonumber 
\alpha^{\ast}=\left(R+\frac{T}{\mu_{2}}S+\frac{1}{c_{2}}I\right)^{-1}e_{1},
\end{align}
where $R=A(B^{T}B)^{-1}A^{T}$, $S_{t}=A_{t}(B_{t}^{T}B_{t})^{-1}A^{T}_{t}$, 
$S=blkdiag(S_{1},\ldots,S_{T})$ and $\alpha^{\ast}=[\alpha_{1}^{\ast^{T}},\ldots,\alpha_{T}^{\ast^{T}}]^{T}$.
As a result, the classifier parameters $u_{0}, u_{t},v_{0}$ and $v_{t}$ of the $t$-th task are determined.

\section{Multi-task twin support vector machine with Universum data}
Motivated by $\mathfrak{U}$-TSVM and DMTSVM, we would like to introduce a new multi-task model and name it as multi-task twin support vector machine with Universum data ($\mathfrak{U}$MTSVM).

For a multi-task problem, we have $\widetilde{T}$ training sets, and suppose that the training set $\widetilde{T}_{t}$ for $t=1,\ldots,T$ consists of two subsets and each task $t$ contains $n_{t}$ samples as follows
	$$\widetilde{T}_{t}=S_{t}\cup X_{\mathfrak{U}t},$$
	where
	\begin{align*}
	& S_{t}=\{(x_{1t}, y_{1t}),\ldots,(x_{{n_{t}}t}, y_{{n_{t}}t}) \},
   \\
	& X_{\mathfrak{U}t}=\{x_{1t}^{\ast},\ldots,x_{u_{t}t}^{\ast} \},
	\end{align*}
with $x_{it}\in \mathbb{R}^{n}$, $y_{i}\in \{\pm 1\}$ and $i=1,\ldots,n_{t}$. 
Hence, the set $S_{t}$ denotes the labeled
samples for the $t$-th task, and  $X_{\mathfrak{U}t}$ contains the Universum data for the $t$-th task. For every task, we expect to build the classifier based on positive  and negative labeled samples as well as Universum data of this task.

 All tasks have two mean hyperplanes  $u_{0}=[w_{1},b_{1}]^{T}$  and $v_{0}=[w_{2},b_{2}]^{T}$.                  
The two  hyperplanes in the $t$-th task for positive and negative classes are $(u_{0}+u_{t})=[w_{1t},b_{1t}]^{T}$  and  $(v_{0}+v_{t})=[w_{2t},b_{2t}]^{T}$, respectively. 
We employ the same notation of $X_{p},X_{pt}, X_{n},X_{nt}, A, A_{t}, B$ and  $B_{t}$  as we used in subsections~\ref{2.1} and~\ref{2.2}. 
In addition, $X_{\mathfrak{U}}$   denotes  the Universum samples, and Universum samples in $t$-th task is presented by matrix~$X_{\mathfrak{U}t}$. Then, for every 
task  $t\in \{1,\ldots,T\}$, we can define:
\[\mathfrak{U}=[X_{\mathfrak{U}}~e_{u}] ~\mbox{ and } ~\mathfrak{U}_t=[X_{\mathfrak{U}t}~e_{ut}],\]
where   $e_{u}$  and   $e_{ut}$  are vector ones of appropriate dimensions.

\subsection{Linear case}
 In this part, we introduce the linear case of our new model ($\mathfrak{U}$MTSVM).  Before define the $ \delta-$insensitive loss function for Universum data, we define the hinge loss function as $H_{\delta}[\theta]=\max \lbrace 0,\delta-\theta \rbrace $.
	 We will define the $ \delta-$insensitive loss $ U^{t}[\theta]=H^{t}_{-\delta}[\theta]+H^{t}_{-\delta}[-\theta],~t=1,...,T $ for Universum data in each task.
	 This loss measures the real-valued output of our classifier 
$ f_{w_{1t},b_{1t}}(x)=w_{1t}^{T}x+b_{1t} $ and $ f_{w_{2t},b_{2t}}(x)=w_{2t}^{T}x+b_{2t} $ on $X_{\mathfrak{U}}$ and penalizes outputs that are far from zero \cite{qi2012twin}. We then wish to minimize the total losses $ \sum_{t=1}^{T} \sum_{j=1}^{u_{t}} U^{t}[f_{w_{1t},b_{1t}}(x^{*}_{jt})] $ and $ \sum_{t=1}^{T} \sum_{j=1}^{u_{t}} U^{t}[f_{w_{2t},b_{2t}}(x^{*}_{jt})]$, and the classifiers have greater possibility when these values are less, and vice versa \cite{zhang2008semi}. Therefore by adding the following  terms in the objective functions of DMTSVM, we introduce our new model ($\mathfrak{U}$MTSVM):
\[c_{u} \sum_{t=1}^{T} \sum_{j=1}^{u_{t}} U^{t}[f_{w_{1t},b_{1t}}(x^{*}_{jt})], ~~\mbox{and} ~~c^{*}_{u}\sum_{t=1}^{T} \sum_{j=1}^{u_{t}} U^{t}[f_{w_{2t},b_{2t}}(x^{*}_{jt})],\]
where $ c_{u}  $ and $ c^{*}_{u} $ controls the loss of  Universum data. Hence,  the optimization formulas of our proposed $\mathfrak{U}$MTSVM  may be stated as follows:


\begin{align}
\mathop {\min }_{u_{0},u_{t},\xi_{t},\psi _{t}\,} \,& \frac{1}{2}\|A u_{0}\|^{2}+\dfrac{\mu_{1}}{2T} \sum_{t=1}^{T}\|A_{t}u_{t}\|^{2}    +c_{1}\sum_{t=1}^{T} e_{2t}^{T} \xi _{t}+c_{u}\sum_{t=1}^{T} e_{ut}^{T} \psi _{t}\nonumber \\
\text{ s.t.} ~~& -  B_{t} (u_{0}+u_{t})+{{\xi }_{t}}\geq e_{2t},\nonumber \\ 
& \mathfrak{U}_{t} (u_{0}+u_{t})+\psi_{t}\geq  (-1+\varepsilon)e_{ut}, \nonumber\\ 
& \xi _{t} \geq 0,  \ \ 
 \psi_{t}\geq 0,
\label{26}
\end{align}
and
\begin{align}
\mathop {\min }_{v_{0},v_{t},\eta_{t},\psi^{*} _{t}\,} \,& \frac{1}{2}\|B v_{0}\|^{2}+\dfrac{\mu_{2}}{2T} \sum_{t=1}^{T}\|B_{t}v_{t}\|^{2}  +c_{2}\sum_{t=1}^{T} e_{1t}^{T} \eta_{t}+c^{*}_{u}\sum_{t=1}^{T} e_{ut}^{T} \psi _{t}^{*}\nonumber  \\
\text{ s.t.}~~ &   A_{t} (v_{0}+v_{t})+{{\eta }_{t}}\geq e_{1t},\nonumber \\ 
& -\mathfrak{U}_{t} (v_{0}+v_{t})+\psi^{*}_{t}\geq  (-1+\varepsilon)e_{ut}, \nonumber\\ 
& \eta _{t} \geq 0,  \ \  
 \psi^{*}_{t}\geq 0,
\label{27}
\end{align}
where  $c_{1}, c_{2}, c_{u}$ and  $c_{u}^{*}$  are penalty parameters.   $\xi_{t}, \eta_{t}$, $ \psi_{t}=(\psi_{1},...,\psi_{ut})$  and $\psi_{t}^{*}=(\psi_{1}^{*},...,\psi_{ut}^{*})$
are the corresponding slack vectors.  $T$ denotes the number of task parameters, and $\mu_{1}$ and  $\mu_{2}$ are the positive
parameters, which controls preference of the tasks.


The Lagrangian function associated with problem (\ref{26}) is denoted by
\begin{align}\label{28}
L_{1}= & \frac{1}{2}\|A u_{0}\|^{2}+\dfrac{\mu_{1}}{2T} \sum_{t=1}^{T}\|A_{t}u_{t}\|^{2}    +c_{1}\sum_{t=1}^{T} e_{2t}^{T} \xi _{t}
 +c_{u}\sum_{t=1}^{T} e_{ut}^{T} \psi _{t}-\sum_{t=1}^{T} \alpha_{1t}^{T}(-B_{t}(u_{0}+u_{t})+\xi_{t}\nonumber \\
& -e_{2t})- \sum_{t=1}^{T} \beta_{1t}^{T}\xi_{t}-\sum_{t=1}^{T} \alpha_{2t}^{T}(\mathfrak{U}_{t}(u_{0}+u_{t})+\psi_{t}
 -(-1+\varepsilon)e_{ut})- \sum_{t=1}^{T} \beta_{2t}^{T}\psi_{t},
\end{align}
where $\alpha_{1t},\alpha_{2t}, \beta_{1t}$ and $\beta_{2t}$ are the  Lagrange multipliers.
Thus, denoting $\alpha_1=[\alpha_{11}^{T},\ldots, \alpha_{1T}^{T}]^{T}$ and $\alpha_2=[\alpha_{21}^{T},\ldots, \alpha_{2T}^{T}]^{T}$, the KKT necessary and sufficient optimality conditions for \eqref{26} are given by


\begin{align}
\dfrac{\partial L_{1}}{\partial u_{0}}& =A^{T}Au_{0}+B^{T}\alpha_{1}-\mathfrak{U}^{T}\alpha_{2}=0,\label{29} \\
\dfrac{\partial L_{1}}{\partial u_{t}}& =\dfrac{\mu_{1}}{T}A_{t}^{T}A_{t}u_{t}+B_{t}^{T}\alpha_{1t}-\mathfrak{U}_{t}^{T}\alpha_{2t}=0,\label{30} \\
\dfrac{\partial L_{1}}{\partial \xi_{t}}& =c_{1}e_{2}-\alpha_{1}-\beta_{1}=0,\label{31} \\
\dfrac{\partial L_{1}}{\partial \psi_{t}}& =c_{u}e_{u}-\alpha_{2}-\beta_{2}=0.\label{32} 
\end{align}
Since  $\beta_{1}\geq 0$  and  $\beta_{2}\geq 0$, from (\ref{31}) and (\ref{32}), we have 
\begin{align} \nonumber                                                     0\leq \alpha_{1}\leq c_{1}e_{2}, ~~~
0\leq \alpha_{2}\leq c_{u}e_{u}. \nonumber 
\end{align}
Also, from the equations (\ref{29}) and  (\ref{30}), we have 
\begin{align} \nonumber                                                      u_{0}&=-(A^{T}A)^{-1} (B^{T}\alpha_{1}-\mathfrak{U}^{T}\alpha_{2}), \nonumber \\ 
u_{t}&=-\dfrac{T}{\mu_{1}}(A_{t}^{T}A_{t})^{-1} (B_{t}^{T}\alpha_{1t}-\mathfrak{U}_{t}^{T}\alpha_{2t}). \nonumber 
\end{align}
Then, substituting  $u_{0}$  and  $u_{t}$  into (\ref{28}) :
\begin{align*}                                                           L_{1}={}&\dfrac{1}{2}(\alpha^{T}_{1}B -\alpha_{2}^{T}\mathfrak{U})(A^{T}A)^{-1} (B^{T}\alpha_{1}-\mathfrak{U}^{T}\alpha_{2})  \\
&+\frac{T}{ 2\mu_{1}}\sum_{t=1}^{T}(\alpha_{1t}^{T}B_{t}- \alpha_{2t}^{T}\mathfrak{U}_{t}) (A_{t}^{T}A_{t})^{-1}(B_{t}^{T}\alpha_{1t}-\mathfrak{U}_{t}^{T}\alpha_{2t})\\
& -\frac{T}{ \mu_{1}}\sum_{t=1}^{T}\alpha_{1t}^{T}B_{t}  (A_{t}^{T}A_{t})^{-1} (B_{t}^{T}\alpha_{1t}-\mathfrak{U}_{t}^{T}\alpha_{2t})\\
&-\sum_{t=1}^{T}\alpha_{2t}^{T}\mathfrak{U}_{t}(A_{t}^{T}A_{t})^{-1} (B ^{T}\alpha_{1}-\mathfrak{U}^{T}\alpha_{2})\\
& -\dfrac{T}{ \mu_{1}}  \sum_{t=1}^{T}\alpha_{1t}^{T}\mathfrak{U}_{t}(A_{t}^{T}A_{t})^{-1} (B_{t}^{T}\alpha_{1t}-\mathfrak{U}_{t}^{T}\alpha_{2t})\\
&=-\dfrac{1}{2}(\alpha^{T}_{1}B -\alpha_{2}^{T}\mathfrak{U})(A^{T}A)^{-1} (B^{T}\alpha_{1}-\mathfrak{U}^{T}\alpha_{2})\\
&-\dfrac{T}{ 2\mu_{1}}  \sum_{t=1}^{T}(\alpha_{1t}^{T}B_{t}- \alpha_{2t}^{T}\mathfrak{U}_{t}) (A_{t}^{T}A_{t})^{-1} (B_{t}^{T}\alpha_{1t}-\mathfrak{U}_{t}^{T}\alpha_{2t}).
\end{align*}
Defining
\begin{align*}
Q& =\left[ \begin{matrix}
B\\
\mathfrak{U}
\end{matrix}\right] (A^{T}A)^{-1}\left[ \begin{matrix}
B^{T}&\mathfrak{U}^{T}
\end{matrix}\right],\\
P_{t}& =\left[ \begin{matrix}
B_{t}\\
\mathfrak{U}_{t}
\end{matrix}\right] (A^{T}A)^{-1}\left[ \begin{matrix}
B_{t}^{T}&\mathfrak{U}^{T}_{t}
\end{matrix}\right],\\
P&= blkdiag \,(P_{1},\ldots,P_{T}),
\end{align*}
the dual problem of (\ref{26}) may be expressed as 
\begin{align}\label{37}
\mathop {\max }_{\alpha_{1}, \alpha_{2}} \,& -\frac{1}{2}\left[ \alpha_{1}^{T}, \alpha_{2}^{T}\right] \left( Q+\dfrac{T}{\mu_{1}}P\right)     \left[ \begin{matrix}
\alpha_{1}\\
\alpha_{2}
\end{matrix}\right]
+\left[ \alpha_{1}^{T}, \alpha_{2}^{T}\right]  \left[ \begin{matrix}
e_{2}\\
(-1+\varepsilon)e_{u}
\end{matrix}\right]\nonumber\\
\text{s.t.} \,\,\,\,\,& 0\leq\alpha_{1}\leq c_{1}e_{2},\nonumber \\ 
& 0\leq\alpha_{2}\leq c_{u}e_{u}.
\end{align}
Similarly, by introducing 
\begin{align*}
&R =\left[ \begin{matrix}
A\\
\mathfrak{U}
\end{matrix}\right] (B^{T}B)^{-1}\left[ \begin{matrix}
A^{T}&\mathfrak{U}^{T}
\end{matrix}\right], ~~
S_{t} =\left[ \begin{matrix}
A_{t}\\
\mathfrak{U}_{t}
\end{matrix}\right] (A_{t}^{T}A_{t})^{-1}\left[ \begin{matrix}
A_{t}^{T}&\mathfrak{U}^{T}_{t}
\end{matrix}\right],\\
&S= blkdiag\,(S_{1},\ldots, S_{T}),~~
\alpha^{*}_{1}=\big[(\alpha_{11}^{\ast})^T,\ldots,(\alpha_{1T}^{\ast})^T\big]^{T},~~ 
\alpha^{*}_{2}=\big[(\alpha_{21}^{\ast})^T,\ldots,(\alpha_{2T}^{\ast})^T\big]^{T},
\end{align*}
the dual problem of (\ref{27}) can be obtained as 
\begin{align}\label{38}
\mathop {\max }_{\alpha^{*}_{1}, \alpha^{*}_{2}} \,& -\frac{1}{2}\left[ \alpha_{1}^{*T}, \alpha_{2}^{*T}\right] 
\left( R+\dfrac{T}{\mu_{2}}S\right)   
\left[ \begin{matrix}
\alpha^{*}_{1}\\
\alpha^{*}_{2}
\end{matrix}\right]
+\left[ \alpha_{1}^{*T}, \alpha_{2}^{*T}\right]  \left[ \begin{matrix}
e_{1}\\
(-1+\varepsilon)e_{u}
\end{matrix}\right]\nonumber\\
\text{s.t.}\,\,\,\,\, & 0\leq\alpha^{*}_{1}\leq c_{2}e_{1},\nonumber \\ 
& 0\leq\alpha^{*}_{2}\leq c^{*}_{u}e_{u}.
\end{align}
By solving problems (\ref{37}) and (\ref{38}), we find the  $\alpha_{1}, \alpha_{2}, \alpha_{1}^{*}$ and  $\alpha_{2}^{*}$, and then the classifiers parameters  $u_{0}, u_{t}, v_{0}$ and  $v_{t}$   of the  $t$-th  task can be obtained.
The label of a new sample   $x\in \mathbb{R}^{n}$  is  determined by (\ref{15}).
A linear $\mathfrak{U}$MTSVM can be obtained by the steps Algorithm~\ref{A1}.
\begin{algorithm} [t] 
	\caption{\label{A1} A linear  multi-task twin support vector machine with Universum ($\mathfrak{U}$MTSVM) }
	\algsetup{linenosize=\normalsize}
	\renewcommand{\algorithmicrequire}{\textbf{Input:}}
	\begin{algorithmic}[1]
		\normalsize
		\REQUIRE{\mbox{}\\-- The training set $\tilde{T}$ and Universum data $ X_{\mathfrak{U}} $;\\
			-- Decide on the total number of tasks included in the data set and assign this value to T;\\
			-- Select classification task $ S_{t}~(t=1,\ldots,T) $ in training  data set $\tilde{T}$;\\
			-- Divide Universum data $X_{\mathfrak{U}} $ by $t$-task and get $ X_{\mathfrak{U}t}~ (t=1,\ldots,T)$;\\
			-- Choose appropriate parameters
			$c_{1}, c_{2},c_{u}$, $c_{u}^{*}$, $ \mu_{1} $, $ \mu_{2} $, and  parameter $\varepsilon \in (0,1)$.}\\
		{\textbf{The outputs:}}
		\begin{list}{--}{}
			\item $ u_{0},~u_{t},~v_{0}$, and $v_{t}. $
		\end{list}
		
		{\textbf{The process:}}
		
		\STATE
		Solve the optimization problems (16) and (17), and get $ \alpha_{1},~\alpha_{2},~\alpha^{*}_{1}$, and $\alpha^{*}_{2}. $
		\STATE
		Calculate  $ u_{0},~u_{t},~v_{0}$, and $v_{t}. $
		\STATE
		By utilizing the decision function (\ref{15}), assign a new point $ x $ in the $t$-th task to class $ +1 $ or $ -1 $.
		\end{algorithmic}
\end{algorithm}

\subsection{Nonlinear case}

It is understandable that a linear classifier would not be appropriate for training data that are linearly inseparable. To deal with such issues, employ the kernel technique. To that end, we introduce the kernel function $K(\cdot,\cdot)$ and define
 $D=\left[ A_{1}^{T}, B_{1}^{T}, A_{2}^{T}, B_{2}^{T},\ldots, A_{T}^{T},B_{T}^{T}\right] ^{T}$, $\overline{A}=\left[ K(A, D^{T}),e_{1}\right] $, $\overline{A}_{t}=\left[ K(A_{t}, D^{T}),e_{1t}\right] $, $\overline{B}=[K(B, D^{T}),e_{2}]$, $\overline{B}_{t}=\left[ K(B_{t}, D^{T}),e_{2t}\right] $, $\overline{\mathfrak{U}}=\left[ K(X_{\mathfrak{U}}, D^{T}),e_{u}\right] $ and $\overline{\mathfrak{U}}_{t}=\left[ K(X_{\mathfrak{U}t}, D^{T}),e_{ut}\right] $. Hence, the nonlinear formulations for   $\mathfrak{U}$MTSVM  are defined as follow:
\begin{align}
\mathop {\min }_{u_{0},u_{t},\xi_{t},\psi _{t}\,} \,& \frac{1}{2}\|\bar{A} u_{0}\|^{2}+\frac{\mu_{1}}{2T} \sum_{t=1}^{T}\|\bar{A}_{t}u_{t}\|^{2}    +c_{1}\sum_{t=1}^{T} e_{2t}^{T} \xi _{t}
+c_{u}\sum_{t=1}^{T} e_{ut}^{T} \psi _{t}\nonumber \\
\text{s.t.} ~~& -  \overline{B}_{t} (u_{0}+u_{t})+{{\xi }_{t}}\geq e_{2t},\nonumber \\ 
& \bar{\mathfrak{U}}_{t} (u_{0}+u_{t})+\psi_{t}\geq  (-1+\varepsilon)e_{ut}, \nonumber\\ 
& \xi _{t} \geq 0, \ \ 
  \psi_{t}\geq 0,
\label{40}
\end{align}
and
\begin{align}
\mathop {\min }_{v_{0},v_{t},\eta_{t},\psi^{*} _{t}\,} \,& \frac{1}{2}\|\bar{B} v_{0}\|^{2}+\frac{\mu_{2}}{2T} \sum_{t=1}^{T}\|\bar{B}_{t}v_{t}\|^{2}    +c_{2}\sum_{t=1}^{T} e_{1t}^{T} \eta _{t}
+c^{*}_{u}\sum_{t=1}^{T} e_{ut}^{T} \psi _{t}\nonumber  \\
\text{s.t.}  ~~&   \bar{A}_{t} (v_{0}+v_{t})+{{\eta }_{t}}\geq e_{1t},\nonumber \\ 
& -\overline{\mathfrak{U}}_{t} (v_{0}+v_{t})+\psi^{*}_{t}\geq  (-1+\varepsilon)e_{ut}, \nonumber\\ 
& \eta _{t} \geq 0, \ \ 
  \psi^{*}_{t}\geq 0,
\label{41}
\end{align}
here $c_{1}$, $c_{2}$, $c_{u}$ and  $c_{u}^{*}$   are penalty parameters, and
$\xi_{t}$, $\eta_{t}$, $\psi_{t}$  and $\psi_{t}^{*}$
are the corresponding slack vectors. By $T$ we denote the number of task parameters, and $\mu_{1}$ and  $\mu_{2}$ are the positive
parameters, which control preference of the tasks.
After using the Lagrange multipliers and KKT conditions, the duals of problems (\ref{40}) and (\ref{41}) read as follows:
\begin{align}\label{42}
\mathop {\max }_{\alpha_{1}, \alpha_{2}} \,& -\frac{1}{2}\left[ \alpha_{1}^{T}, \alpha_{2}^{T}\right]
\left( Q+\dfrac{T}{\mu_{1}}P\right)     \left[ \begin{matrix}
\alpha_{1}\\
\alpha_{2}
\end{matrix}\right]
+\left[ \alpha_{1}^{T}, \alpha_{2}^{T}\right]  \left[ \begin{matrix}
e_{2}\\
(-1+\varepsilon)e_{u}
\end{matrix}\right]\nonumber \\
\text{s.t.} \,\,\,\,\, &  0\leq\alpha_{1}\leq c_{1}e_{2},\nonumber \\ 
& 0\leq\alpha_{2}\leq c_{u}e_{u},
\end{align}
and
\begin{align}\label{43}
\mathop {\max }_{\alpha^{*}_{1}, \alpha^{*}_{2}} \,& -\frac{1}{2}\left[ \alpha_{1}^{*T}, \alpha_{2}^{*T}\right] 
\left( R+\dfrac{T}{\mu_{2}}S\right)     \left[ \begin{matrix}
\alpha^{*}_{1}\\
\alpha^{*}_{2}
\end{matrix}\right]
+\left[ \alpha_{1}^{*T}, \alpha_{2}^{*T}\right]  \left[ \begin{matrix}
e_{1}\\
(-1+\varepsilon)e_{u}
\end{matrix}\right]\nonumber\\
\text{s.t.}\,\,\,\,\, & 0\leq\alpha^{*}_{1}\leq c_{2}e_{1},\nonumber \\ 
& 0\leq\alpha^{*}_{2}\leq c^{*}_{u}e_{u},
\end{align}
where
\begin{align*}
Q& =\left[ \begin{matrix}
\bar{B}\\
\bar{ \mathfrak{U}}
\end{matrix}\right] ( \bar{A}^{T} \bar{A})^{-1}\left[ \begin{matrix}
\bar{B}^{T}& \bar{ \mathfrak{U}}^{T}
\end{matrix}\right],~~
P_{t} =\left[ \begin{matrix}
\bar{B}_{t}\\
\bar{ \mathfrak{U}}_{t}
\end{matrix}\right] ( \bar{A}_{t}^{T} \bar{A}_{t})^{-1}\left[ \begin{matrix}
\bar{B}_{t}^{T}& \bar{ \mathfrak{U}}^{T}_{t}
\end{matrix}\right],\\
P&=blkdiag \,(P_{1},\ldots, P_{T}),~~
R =\left[ \begin{matrix}
\bar{A}\\
\bar{ \mathfrak{U}}
\end{matrix}\right] (\bar{B}^{T}\bar{B})^{-1}\left[ \begin{matrix}
\bar{A}^{T}& \bar{ \mathfrak{U}}^{T}
\end{matrix}\right],\\
S_{t}& =\left[ \begin{matrix}
\bar{A}_{t}\\
\bar{ \mathfrak{U}}_{t}
\end{matrix}\right] (\bar{A}_{t}^{T}\bar{A}_{t})^{-1}\left[ \begin{matrix}
\bar{A}_{t}^{T}& \bar{ \mathfrak{U}}^{T}_{t}
\end{matrix}\right], ~~
S= blkdiag \,(S_{1},\ldots, S_{T}).
\end{align*}

A new data point $x$ in the $t$-th  task is determined to class $i\in \{+1, -1\}$ by using the following decision function:
\begin{align}\label{n15} 
f(x)=\arg \underset{k=1,2}\min\, \big|K\big(x,D^{T}\big)w_{kt}+b_{kt}\big|.
\end{align}
The nonlinear $\mathfrak{U}$MTSVM is described in the steps of Algorithm~\ref{A2}.
\begin{algorithm} [t] 
	\caption{\label{A2} A nonlinear  multi-task twin support vector machine with Universum ($\mathfrak{U}$MTSVM) }
	\algsetup{linenosize=\normalsize}
	\renewcommand{\algorithmicrequire}{\textbf{Input:}}
	\begin{algorithmic}[1]
		\normalsize
		\REQUIRE{\mbox{}\\-- The training set $\tilde{T}$ and Universum data $X_{\mathfrak{U}} $;\\
			-- Decide on the total number of tasks included in the data set and assign this value to T;\\
			-- Select classification task $ S_{t}~(t=1,\ldots,T) $ in training  data set $\tilde{T}$;\\
			-- Divide Universum data $X_{\mathfrak{U}} $ by $t$-task and get $ X_{\mathfrak{U}t}~ (t=1,\ldots,T)$;\\
			-- Choose appropriate parameters
			$c_{1}, c_{2},c_{u}$, $c_{u}^{*}$, $ \mu_{1} $, $ \mu_{2} $, and  parameter $\varepsilon \in (0,1)$.\\
		-- Select proper kernel function and kernel parameter.
	}\\
		{\textbf{The outputs:}}
		\begin{list}{--}{}
			\item $ u_{0},~u_{t},~v_{0}$, and $v_{t}. $
		\end{list}
		
		{\textbf{The process:}}
		
		\STATE
		Solve the optimization problems (20) and (21), and get $ \alpha_{1},~\alpha_{2},~\alpha^{*}_{1}$, and $\alpha^{*}_{2}. $
		\STATE
		Calculate  $ u_{0},~u_{t},~v_{0}$, and $v_{t}. $
		\STATE
		Assign a new point $ x $ in the $t$-th task to class $ +1 $ or $ -1 $ by using decision function~(\ref{n15}).
	\end{algorithmic}
\end{algorithm}

\section{Least squares multi-task twin support vector machine with Universum data}
In this section, we introduce the least-squares version of $\mathfrak{U}$MTSVM for the linear and nonlinear cases, to which we refer as  least-squares multi-task twin support vector machine with Universum data (LS-$\mathfrak{U}$MTSVM). Our proposed method combines the advantages of DMTSVM, $ \mathfrak{U}_{LS} $-TSVM, and MTLS-TWSVM. In terms of generalization performance, the proposed method is superior to MTLS-TWSVM, because it improves prediction accuracy by absorbing previously embedded knowledge embedded in the Universum data. In terms of the time of computation, LS-$\mathfrak{U}$MTSVM works faster than DMTSVM by solving two systems of linear equations instead of two quadratic programming problems.

\subsection{Linear case}
We modify problems (\ref{26}) and 
(\ref{27}) in the least  squares  sense and  replace the inequality constraint with equality requirements as follows
\begin{align}\label{44}
\mathop {\min }_{u_{0},u_{t},\xi_{t},\psi _{t}\,} \,& \frac{1}{2}\|A u_{0}\|^{2}+\dfrac{\mu_{1}}{2T} \sum_{t=1}^{T}\|A_{t}u_{t}\|^{2}    +\dfrac{c_{1}}{2}\sum_{t=1}^{T} \| \xi _{t}\|^{2}
+\dfrac{c_{u}}{2}\sum_{t=1}^{T} \|\psi _{t}\|^{2}\nonumber \\
\text{s.t.} ~~& -  B_{t} (u_{0}+u_{t})+{{\xi }_{t}}= e_{2t},\nonumber \\ 
& \mathfrak{U}_{t} (u_{0}+u_{t})+\psi_{t}=  (-1+\varepsilon)e_{ut}, 
\end{align}
and
\begin{align}\label{45}
\mathop {\min }_{v_{0},v_{t},\eta_{t},\psi^{*} _{t}\,} \,& \frac{1}{2}\|B v_{0}\|^{2}+\dfrac{\mu_{2}}{2T} \sum_{t=1}^{T}\|B_{t}v_{t}\|^{2}    +\dfrac{c_{2}}{2}\sum_{t=1}^{T} \| \eta_{t}\|^{2}
+\dfrac{c^{*}_{u}}{2}\sum_{t=1}^{T}\|\psi _{t}^{*}\|^{2}\nonumber  \\
\text{s.t.}~~ &   A_{t} (v_{0}+v_{t})+{{\eta }_{t}}= e_{1t},\nonumber \\ 
& -\mathfrak{U}_{t} (v_{0}+v_{t})+\psi^{*}_{t}=  (-1+\varepsilon)e_{ut}.
\end{align}
Here, $c_{1},c_{2},c_{u}$,  and   $c_{u}^{\ast}$   are penalty parameters,   $\xi_{t}, \eta_{t}, \psi_{t}$  and  $\psi_{t}^{\ast}$   are slack variables for $t$-th task and  $e_{1t}, e_{2t}$,  and  $e_{ut}$  are vectors of appropriate dimensions whose all components are equal  to $1$.

It is worth noting that the loss functions in \eqref{44} and \eqref{45} are the square of the 2-norm of the slack variables $\psi$ and $\psi^{*}$ rather than the 1-norm  in problems (\ref{26}) and (\ref{27}), which renders the constraints $\psi_{t}\geq 0$ and $\psi^{*}_{t}\geq 0$ superfluous.

The Lagrangian function for the problem \eqref{44} can be written as follows: 
\begin{align}\label{46} 
L_{1} ={}& \dfrac{1}{2}\| Au_{0}\|^{2}+\frac{\mu_{1}}{2T}\sum_{t=1}^{T} \| A_{t}u_{t}\|^{2}+\frac{c_{1}}{2}\sum_{t=1}^{T}\| \xi_{t}\|^{2}\nonumber\\
& -\sum_{t=1}^{T}\alpha_{t}^{T}(- B_{t}(u_{0}+u_{t})+\xi_{t}-e_{2t})+\dfrac{c_{u}}{2}\sum_{t=1}^{T}\| \psi_{t}\|^{2}\nonumber\\
& -\sum_{t=1}^{T}\beta_{t}^{T}(\mathfrak{U}_{t}(u_{0}+u_{t})+\psi_{t}-(-1+\varepsilon)e_{ut}),
\end{align}
where  $\alpha_{t}$   and  $\beta_{t}$   are the Lagrange multipliers.
The Lagrangian function \eqref{46} is differentiable and the KKT optimally conditions can be obtained as follows:
\begin{align}
&\dfrac{\partial L}{\partial u_{0}}=A^{T}Au_{0}+B^{T}\alpha -\mathfrak{U}^{T} \beta =0,\label{47} \\
&\dfrac{\partial L}{\partial u_{t}}=\dfrac{\mu_{1}}{T}A^{T}_{t}A_{t}u_{t}+B^{T}_{t}\alpha_{t} - \mathfrak{U}^{T}_{t} \beta_{t}=0,\label{48} \\
&\dfrac{\partial L}{\partial \xi_{t}}=c_{1}\xi_{t}-\alpha_{t} =0, \label{49} \\
&\dfrac{\partial L}{\partial \psi_{t}}=c_{u}\psi_{t}-\beta_{t} =0, \label{50} \\
&\dfrac{\partial L_{1}}{\partial \alpha_{t}}=B_{t}(u_{0}+u_{t})-\xi_{t}+e_{2t}=0,\label{51} \\
&\dfrac{\partial L_{1}}{\partial \beta_{t}}=-\mathfrak{U}_{t}(u_{0}+u_{t})-\psi_{t}+(-1+\varepsilon)e_{ut}=0.\label{52} 
\end{align}
From equations \eqref{47}--\eqref{50}, we derive
\begin{align}
& u_{0}= -(A^{T}A)^{-1}(B^{T}\alpha -\mathfrak{U}\beta), \label{53}\\ 
& u_{t}= -\dfrac{T}{\mu_{1}}(A^{T}_{t}A_{t})^{-1}(B^{T}_{t}\alpha_{t}-\mathfrak{U}_{t}^{T}\beta_{t}), \label{54}\\ 
& \xi_{t}=\dfrac{\alpha_{t}}{c_{1}},\label{55} \\
& \psi_{t}=\dfrac{\beta_{t}}{c_{u}}.\label{56} 
\end{align}
By substituting  $u_{0}$,  $u_{t}$,  $\xi_{t}$  and  $\psi_{t}$  into the equations  \eqref{51} and \eqref{52}, we have 
\begin{align}\label{57} 
{{B}_{t}}\left[ -{{({{A}^{T}}A)}^{-1}}({B}^{T}\alpha - \mathfrak{U}^{T}\beta)
-\frac{T}{{{\mu }_{1}}}{{({{A}^{T}_{t}}A_{t})}^{-1}}(B_{t}^{T}{{\alpha }_{t}}
-\mathfrak{U}_{t}^{T}\beta_{t}) \right]
-\frac{{{\alpha }_{t}}}{{{c}_{1}}}
&=  -{{e}_{2t}},
\\
\label{58} 
{ -\mathfrak{U}_{t}}\left[ -{{({{A}^{T}}A)}^{-1}}({B}^{T}\alpha - \mathfrak{U}^{T}\beta)
 -\frac{T}{{{\mu }_{1}}}{{({{A}^{T}_{t}}A_{t})}^{-1}}(B_{t}^{T}{{\alpha }_{t}}-\mathfrak{U}_{t}^{T}\beta_{t}) \right]-\frac{{{\beta }_{t}}}{{{c}_{u}}}
 &=-(-1 + \varepsilon){{e}_{ut}},
\end{align}
where   $t\in \{1, \ldots, T\}$, $\alpha=[\alpha_{1}^{T},\ldots, \alpha_{T}^{T}]^{T}$  and $\beta=[\beta_{1}^{T},\ldots, \beta_{T}^{T}]^{T}$. Here, we define 
\begin{align*}
& Q_{1} =B{{({{A}^{T}}A)}^{-1}}{{B}^{T}}, ~~
 Q_{2} =B{{({{A}^{T}}A)}^{-1}}{{\mathfrak{U}}^{T}}, ~~
S_{1} =\mathfrak{U}{{({{A}^{T}}A)}^{-1}}{{B}^{T}}, ~~ S_{2} =\mathfrak{U}{{({{A}^{T}}A)}^{-1}}{{\mathfrak{U}}^{T}}, 
\\
& {{P}_{1t}}={{B}_{t}}{{(A_{t}^{T}A_{t})}^{-1}}B_{t}^{T}, ~~ {{P}_{2t}}={{B}_{t}}{{(A_{t}^{T}A_{t})}^{-1}}\mathfrak{U}_{t}^{T},~~ P_{1}=blkdiag({{P}_{11}},\ldots ,{{P}_{1T}}),
\\
& P_{2}=blkdiag({{P}_{21}},\ldots ,{{P}_{2T}}),~~ R_{1t}=\mathfrak{U}_{t}{{({{A}^{T}_{t}}A_{t})}^{-1}}{{B}^{T}_{t}},~~ R_{2t}=\mathfrak{U}_{t}{{({{A}^{T}_{t}}A_{t})}^{-1}}{{\mathfrak{U}}^{T}_{t}}, 
\\
& R_{1}=blkdiag({{R}_{11}},\ldots ,{{R}_{1T}}),~~ R_{2}=blkdiag({{R}_{21}},\ldots ,{{R}_{2T}}).
\end{align*}
Then \eqref{57} and \eqref{58} can be converted to the following equations:
\begin{align}
& \left( {{Q}_{1}}\alpha -{{Q }_{2}}\beta  \right)+\frac{T}{{{\mu }_{1}}}\left( {{P}_{1}}\alpha -{{P}_{2}}\beta  \right)+\frac{1}{{{c}_{1}}}{{I}_{1}}\alpha  ={{e}_{2}}, \label{59} \\ 
& \left( -{{S}_{1}}\alpha +{{S}_{2}}\beta  \right)-\frac{T}{{{\mu }_{2}}}\left( {{R}_{1}}\alpha -{{R}_{2}}\beta  \right)+\frac{1}{{{c}_{u}}}{{I}_{2}}\beta 
=\left( -1+\varepsilon  \right){{e}_{u}}. \label{60}
\end{align}
Combining equations \eqref{59} and\eqref{60}, we obtain
\begin{align}
\left[ \begin{matrix}
Q_{1}&-Q_{2}\\
-S_{1}&S_{2}
\end{matrix}\right] \left[\begin{matrix}
\alpha\\
\beta
\end{matrix} \right] &+ \dfrac{T}{\mu_{1}}\left[ \begin{matrix}
P_{1}&-P_{2}\\
-R_{1}&R_{2}
\end{matrix}\right] \left[\begin{matrix}
\alpha\\
\beta
\end{matrix} \right] 
+ \left[ \begin{matrix}
\dfrac{1}{c_{1}}I_{1}&0\\
0&\dfrac{1}{c_{u}}I_{2}
\end{matrix}\right] \left[\begin{matrix}
\alpha\\
\\
\beta
\end{matrix} \right] 
= \left[\begin{matrix}
e_{2}\\
\\
(-1+\varepsilon)e_{u}
\end{matrix} \right],\label{61} 
\end{align}
then, we can write
\begin{align}
\left[ \begin{matrix}
\alpha   \\
\beta   \\
\end{matrix} \right]&=\left[ \left[ \begin{matrix}
{{Q}_{1}} & -{{Q }_{2}}  \\
-{{S}_{1}} & {{S}_{2}}  \\
\end{matrix} \right]+\frac{T}{{{\mu }_{1}}}\left[ \begin{matrix}
{{P}_{1}} & -{{P}_{2}}  \\
-{{R}_{1}} & {{R}_{2}}  \\
\end{matrix} \right] 
 + \left[ \begin{matrix}
\dfrac{1}{{{c}_{1}}}{{I}_{1}} & 0  \\
0 & \dfrac{1}{{{c}_{u}}}{{I}_{2}}  \\
\end{matrix} \right] \right]^{-1}\left[ \begin{matrix}
{{e}_{2}}  \\
(-1+\varepsilon ){{e}_{u}}  
\end{matrix} \right]. \label{62} 
\end{align}
Similarly, the Lagrangian function for the problem \eqref{45} can be written as follows:
\begin{align}\label{63} 
L_{2} ={}& \dfrac{1}{2}\| Bv_{0}\|^{2}+\dfrac{\mu_{2}}{2T}\sum_{t=1}^{T} \| B_{t}v_{t}\|^{2}+\dfrac{c_{2}}{2}\sum_{t=1}^{T}\| \eta_{t}\|^{2}\nonumber\\
& -\sum_{t=1}^{T}\alpha_{t}^{\ast^{T}}(A_{t}(v_{0}+v_{t})+\eta_{t}-e_{1t})+\dfrac{c_{u}^{\ast}}{2}\sum_{t=1}^{T}\| \psi_{t}^{\ast}\|^{2}\nonumber\\
& -\sum_{t=1}^{T}\beta_{t}^{\ast^{T}}(\mathfrak{U}_{t}(v_{0}+v_{t})+\psi_{t}^{\ast}-(-1+\varepsilon)e_{ut}).
\end{align}
By performing a similar process, the following equations are obtained:
\begin{align}
	(Q _{1}^{*}\alpha _{1}^{*}-Q _{2}^{*}{{\beta }^{*}})+\frac{T}{{{\mu }_{2}}}\left( P_{1}^{*}{{\alpha }^{*}}-P_{2}^{*}{{\beta }^{*}} \right)+\frac{1}{{{c}_{2}}}{{I}_{1}}{{\alpha }^{*}}
&={{e}_{1}},\label{64} 
\\
\left( {{S}^{*}_{1}}\alpha^{*}-S_{2}^{*}{{\beta }^{*}} \right)+\frac{T}{{{\mu }_{2}}}( R_{1}^{*}{{\alpha }^{*}}
-R_{2}^{*}{{\beta }^{*}} )
-\frac{1}{{{c}_{u}^{\ast}}}{{I}_{2}}{{\beta }^{*}}
&= -(-1+\varepsilon)e_{u}, \label{65} 
\end{align}
where
\begin{align*}
& Q_{1}^{\ast} =A{{({{B}^{T}}B)}^{-1}}{{A}^{T}}, ~~ Q_{2}^{\ast} =A{{({{B}^{T}}B)}^{-1}}{{\mathfrak{U}}^{T}}, ~~ S_{1}^{\ast} =\mathfrak{U}{{({{B}^{T}}B)}^{-1}}{{A}^{T}},  
\\
&S_{2}^{\ast} =\mathfrak{U}{{({{B}^{T}}B)}^{-1}}{{\mathfrak{U}}^{T}},~~ {{P}_{1t}^{\ast}}={{A}_{t}}{{(B_{t}^{T}B_{t})}^{-1}}A_{t}^{T}, ~~ {{P}_{2t}^{\ast}}={{A}_{t}}{{(B_{t}^{T}B_{t})}^{-1}}\mathfrak{U}_{t}^{T},\\
& P_{1}^{\ast}=blkdiag({{P}_{11}^{\ast}},\ldots ,{{P}_{1T}^{\ast}}),~~ P_{2}^{\ast}=blkdiag({{P}_{21}^{\ast}},\ldots ,{{P}_{2T}^{\ast}}),~~ R_{1t}^{\ast}=\mathfrak{U}_{t}{{({{B}^{T}_{t}}B_{t})}^{-1}}{{A}^{T}_{t}}, \\
& R_{2t}^{\ast}=\mathfrak{U}_{t}{{({{B}^{T}_{t}}B_{t})}^{-1}}{{\mathfrak{U}}^{T}_{t}}, 
~~ R_{1}^{\ast}=blkdiag({{R}_{11}^{\ast}},\ldots ,{{R}_{1T}^{\ast}}),
~~ R_{2}^{\ast}=blkdiag({{R}_{21}^{\ast}},\ldots ,{{R}_{2T}^{\ast}}).
\end{align*}
Combining equations \eqref{64} and \eqref{65}, we have
\begin{align}
\left[ \begin{matrix}
Q_{1}^{\ast}&-Q_{2}^{\ast}\\
S_{1}^{\ast}&-S_{2}^{\ast}
\end{matrix}\right] \left[\begin{matrix}
\alpha^{\ast}\\
\beta^{\ast}
\end{matrix} \right] &+ \dfrac{T}{\mu_{2}}\left[ \begin{matrix}
P_{1}^{\ast}&-P_{2}^{\ast}\\
R_{1}^{\ast}&-R_{2}^{\ast}
\end{matrix}\right] \left[\begin{matrix}
\alpha^{\ast}\\
\beta^{\ast}
\end{matrix} \right] 
+ \left[ \begin{matrix}
\dfrac{1}{c_{2}}I_{1}&0\\
0&-\dfrac{1}{c_{u}^{\ast}}I_{2}
\end{matrix}\right] \left[\begin{matrix}
\alpha^{\ast}\\
\\
\beta^{\ast}
\end{matrix} \right] 
= \left[\begin{matrix}
e_{1}\\
\\
-(-1+\varepsilon)e_{u}
\end{matrix} \right],
\end{align}
then, we can write
\begin{align}
\left[ \begin{matrix}
\alpha^{\ast}   \\
\beta^{\ast}   \\
\end{matrix} \right]&=\left[ \left[ \begin{matrix}
{{Q}_{1}^{\ast}} & -{{Q }_{2}^{\ast}}  \\
{{S}_{1}^{\ast}} & {-{S}_{2}^{\ast}}  \\
\end{matrix} \right]+\frac{T}{{{\mu }_{2}}}\left[ \begin{matrix}
{{P}_{1}^{\ast}} & -{{P}_{2}^{\ast}}  \\
{{R}_{1}^{\ast}} & -{{R}_{2}^{\ast}}  \\
\end{matrix} \right] + \left[ \begin{matrix}
\dfrac{1}{{{c}_{2}}}{{I}_{1}} & 0  \\
0 & \dfrac{1}{{{c}_{u}^{\ast}}}{{I}_{2}}  \\
\end{matrix} \right] \right]^{-1}\left[ \begin{matrix}
{{e}_{1}}  \\
- (-1+\varepsilon ){{e}_{u}}  
\end{matrix} \right].\label{66} 
\end{align}
Finally, by finding  solutions \eqref{62} and \eqref{66}, the classifier parameters  $u_{0}$,  $u_{t}$,  $v_{0}$   and   $v_{t}$  are obtained.
The decision function \eqref{15}  can be used to assign a new data point  $x\in \mathbb{R}^{n}$   to its appropriate class.
According to the discussion above, we illustrate the LS-$\mathfrak{U}$MTSVM via Algorithm~\ref{A3}.
\begin{algorithm} [t] 
	\caption{\label{A3} A linear least squares multi-task twin support vector machine with Universum (LS-$\mathfrak{U}$MTSVM) }
	\algsetup{linenosize=\normalsize}
	\renewcommand{\algorithmicrequire}{\textbf{Input:}}
	\begin{algorithmic}[1]
		\normalsize
		\REQUIRE{\mbox{}\\-- The training set $\tilde{T}$ and Universum data $ X_{\mathfrak{U}} $;\\
			-- Decide on the total number of tasks included in the data set and assign this value to T;\\
			-- Select classification task $ S_{t}~(t=1,\ldots,T) $ in training  data set $\tilde{T}$;\\
			-- Divide Universum data $ X_{\mathfrak{U}} $ by $t$-task and get $ X_{\mathfrak{U}t}~ (t=1,\ldots,T)$;\\
			-- Choose appropriate parameters
			$c_{1}, c_{2},c_{u}$, $c_{u}^{*}$, $ \mu_{1} $, $ \mu_{2} $, and  parameter $\varepsilon \in (0,1)$.}\\
		{\textbf{The outputs:}}
		\begin{list}{--}{}
			\item $ u_{0},~u_{t},~v_{0}$, and $v_{t}. $
		\end{list}
		
		{\textbf{The process:}}
		
		\STATE
		Solve the two small systems of linear equations (41) and (46), and get $ \alpha,~\beta,~\alpha^{*}$, and $\beta^{*}. $
		\STATE
		Calculate  $ u_{0},~u_{t},~v_{0}$, and $v_{t}. $
		\STATE
			By utilizing the decision function (\ref{15}), assign a new point $ x $ in the $t$-th task to class $ +1 $ or $ -1 $.
	\end{algorithmic}
\end{algorithm}

\subsection{Nonlinear case}

In the following, we introduce a nonlinear version of  our proposed LS-$\mathfrak{U}$MTSVM  because there are situations that are not linearly separable, in which case the kernel trick can be used.
Therefore, we use the kernel function $K(\cdot,\cdot)$ and define
\begin{align*}
& D={{\left[ A_{1}^{T},B_{1}^{T},A_{2}^{T},B_{2}^{T},\ldots ,A_{T}^{T},B_{T}^{T} \right]}^{T}}, \\ 
& \overline{A}=\left[ K(A,{{D}^{T}}),e_{1} \right],\,\,\,\,{{\overline{A}}_{t}}=\left[ K({{A}_{t}},{{D}^{T}}),{{e}_{1t}} \right], \\ 
& \overline{B}=\left[ K(B,{{D}^{T}}),e_{2} \right],\,\,\,\,{{\overline{B}}_{t}}=\left[ K({{B}_{t}},{{D}^{T}}),{{e}_{2t}} \right], \\ 
& \overline{\mathfrak{U}}=\left[ K(X_{\mathfrak{U}},{{D}^{T}}),e_{u} \right],\,\,\,\,{{\overline{\mathfrak{U}}}_{t}}=\left[ K({X_{\mathfrak{U}t}},{{D}^{T}}),{{e}_{ut}} \right].
\end{align*}
So, the nonlinear formulations of the optimization problems \eqref{44} and \eqref{45} can be written as
\begin{align}\label{67} 
\underset{u_{0}, u_{t}, \xi_{t},\psi_{t}}\min  ~& \dfrac{1}{2}\| \overline{A}u_{0}\|^{2} +\dfrac{\mu_{1}}{2T}\sum_{t=1}^{T} \| \overline{A}_{t}u_{t}\|^{2}+\dfrac{c_{1}}{2}\sum_{t=1}^{T}\| \xi_{t}\|^{2}+\dfrac{c_{u}}{2}\sum_{t=1}^{T}\| \psi_{t}\|^{2}\nonumber\\
\text{s.t.}\,\,\,\,\,\,& -\overline{B}_{t}(u_{0}+u_{t})+\xi_{t}= e_{2t},\\
&\overline{\mathfrak{U}}_{t}(u_{0}+u_{t})+\psi_{t}=(-1+\varepsilon)e_{ut},\nonumber
\end{align}
and
\begin{align}\label{68} 
\underset{v_{0}, v_{t}, \eta_{t},\psi_{t}^{\ast}}\min ~&\dfrac{1}{2}\| \overline{B}v_{0}\|^{2} +\dfrac{\mu_{2}}{2T}\sum_{t=1}^{T} \| \overline{B}_{t}v_{t}\|^{2}+\dfrac{c_{2}}{2}\sum_{t=1}^{T}\| \eta_{t}\|^{2}+\dfrac{c_{u}^{\ast}}{2}\sum_{t=1}^{T}\| \psi_{t}^{\ast}\|^{2}\nonumber\\
\text{s.t.}\,\,\,\,\,\,& \overline{A}_{t}(v_{0}+v_{t})+\eta_{t}= e_{1t},\\
&-\overline{\mathfrak{U}}_{t}(v_{0} +v_{t})+\psi_{t}^{\ast}=(-1+\varepsilon)e_{ut},\nonumber
\end{align}
here, parameters $ c_{1}$, $c_{2}$, $c_{u}$, $c_{u^*}$, $\mu_{1}$, and $\mu_{2} $ are as defined in section~4.1. In a similar way to the linear case, we can written the Lagrangian function problems (\ref{67}) and (\ref{68}) and KKT optimality conditions. After that, the optimal solutions of problems (\ref{67}) and (\ref{68}) take the form
\begin{align}
\left[ \begin{matrix}
\alpha   \\
\beta   \\
\end{matrix} \right]
&=\left[ \left[ \begin{matrix}
{{Q}_{1}} & -{{Q }_{2}}  \\
-{{S}_{1}} & {{S}_{2}}  \\
\end{matrix} \right]+\frac{T}{{{\mu }_{1}}}\left[ \begin{matrix}
{{P}_{1}} & -{{P}_{2}}  \\
-{{R}_{1}} & {{R}_{2}}  \\
\end{matrix} \right]   
+ \left[ \begin{matrix}
\dfrac{1}{{{c}_{1}}}{{I}_{1}} & 0  \\
0 & \dfrac{1}{{{c}_{u}}}{{I}_{2}}  \\
\end{matrix} \right] \right]^{-1}\left[ \begin{matrix}
{{e}_{2}}  \\
(-1+\varepsilon ){{e}_{u}}  
\end{matrix} \right], \label{69} 
\end{align}
and
\begin{align}
\left[ \begin{matrix}
\alpha^{\ast}   \\
\beta^{\ast}   \\
\end{matrix} \right]&=\left[ \left[ \begin{matrix}
{{Q}_{1}^{\ast}} & -{{Q }_{2}^{\ast}}  \\
{{S}_{1}^{\ast}} & {-{S}_{2}^{\ast}}  \\
\end{matrix} \right]+\frac{T}{{{\mu }_{2}}}\left[ \begin{matrix}
{{P}_{1}^{\ast}} & -{{P}_{2}^{\ast}}  \\
{{R}_{1}^{\ast}} & -{{R}_{2}^{\ast}}  \\
\end{matrix} \right]  + \left[ \begin{matrix}
\dfrac{1}{{{c}_{2}}}{{I}_{1}} & 0  \\
0 & \dfrac{1}{{{c}_{u}^{\ast}}}{{I}_{2}}  \\
\end{matrix} \right] \right]^{-1}\left[ \begin{matrix}
{{e}_{1}}  \\
- (-1+\varepsilon ){{e}_{u}}  
\end{matrix} \right],\label{70} 
\end{align}
where
\begin{align*}
& Q_{1} =\overline{B}{{({{\overline{A}}^{T}}\overline{A})}^{-1}}{{\overline{B}}^{T}}, ~~
Q_{2} =\overline{B}{{({{\overline{A}}^{T}}\overline{A})}^{-1}}{{\overline{\mathfrak{U}}}^{T}}, ~~
S_{1} =\overline{\mathfrak{U}}{{({{\overline{A}}^{T}}\overline{A})}^{-1}}{{\overline{B}}^{T}}, ~~ S_{2} =\overline{\mathfrak{U}}{{({{\overline{A}}^{T}}\overline{A})}^{-1}}{{\overline{\mathfrak{U}}}^{T}}, 
\\
& {{P}_{1t}}={{\overline{B}}_{t}}{{(\overline{A}_{t}^{T}\overline{A}_{t})}^{-1}}\overline{B}_{t}^{T}, ~~ {{P}_{2t}}={{\overline{B}}_{t}}{{(\overline{A}_{t}^{T}\overline{A}_{t})}^{-1}}\overline{\mathfrak{U}}_{t}^{T},~~ P_{1}=blkdiag({{P}_{11}},\ldots ,{{P}_{1T}}),
\\
& P_{2}=blkdiag({{P}_{21}},\ldots ,{{P}_{2T}}),~~ R_{1t}=\overline{\mathfrak{U}}_{t}{{({{\overline{A}}^{T}_{t}}\overline{A}_{t})}^{-1}}{{\overline{B}}^{T}_{t}},~~ R_{2t}=\overline{\mathfrak{U}}_{t}{{({{\overline{A}}^{T}_{t}}\overline{A}_{t})}^{-1}}{{\overline{\mathfrak{U}}}^{T}_{t}}, 
\\
& R_{1}=blkdiag({{R}_{11}},\ldots ,{{R}_{1T}}),~~ R_{2}=blkdiag({{R}_{21}},\ldots ,{{R}_{2T}}),
\end{align*}
and
\begin{align*}
& Q_{1}^{\ast} =\overline{A}{{({{\overline{B}}^{T}}\overline{B})}^{-1}}{{\overline{A}}^{T}}, ~~ Q_{2}^{\ast} =\overline{A}{{({{\overline{B}}^{T}}\overline{B})}^{-1}}{{\overline{\mathfrak{U}}}^{T}}, ~~ S_{1}^{\ast} =\overline{\mathfrak{U}}{{({{\overline{B}}^{T}}\overline{B})}^{-1}}{{\overline{A}}^{T}}, 
\\
&S_{2}^{\ast} =\overline{\mathfrak{U}}{{({{\overline{B}}^{T}}\overline{B})}^{-1}}{{\overline{\mathfrak{U}}}^{T}},~~ {{P}_{1t}^{\ast}}={{\overline{A}}_{t}}{{(\overline{B}_{t}^{T}\overline{B}_{t})}^{-1}}\overline{A}_{t}^{T}, ~~ {{P}_{2t}^{\ast}}={{\overline{A}}_{t}}{{(\overline{B}_{t}^{T}\overline{B}_{t})}^{-1}}\overline{\mathfrak{U}}_{t}^{T},
\\
& P_{1}^{\ast}=blkdiag({{P}_{11}^{\ast}},\ldots ,{{P}_{1T}^{\ast}}),~~ P_{2}^{\ast}=blkdiag({{P}_{21}^{\ast}},\ldots ,{{P}_{2T}^{\ast}}),~~ R_{1t}^{\ast}=\overline{\mathfrak{U}}_{t}{{({{\overline{B}}^{T}_{t}}\overline{B}_{t})}^{-1}}{{\overline{A}}^{T}_{t}}, 
\\
& R_{2t}^{\ast}=\overline{\mathfrak{U}}_{t}{{({{\overline{B}}^{T}_{t}}\overline{B}_{t})}^{-1}}{{\overline{\mathfrak{U}}}^{T}_{t}}, 
~~ R_{1}^{\ast}=blkdiag({{R}_{11}^{\ast}},\ldots ,{{R}_{1T}^{\ast}}),
~~ R_{2}^{\ast}=blkdiag({{R}_{21}^{\ast}},\ldots ,{{R}_{2T}^{\ast}}).
\end{align*}
Then the corresponding decision function of the $t$-th task can be computed by~(\ref{n15}). 
Algorithm~\ref{A4} describes the process of nonlinear case.

\begin{algorithm} [t] 
	\caption{\label{A4} A nonlinear least squares multi-task twin support vector machine with Universum (LS-$\mathfrak{U}$MTSVM)}
	\algsetup{linenosize=\normalsize}
	\renewcommand{\algorithmicrequire}{\textbf{Input:}}
	\begin{algorithmic}[1]
		\normalsize
		\REQUIRE{\mbox{}\\-- The training set $\tilde{T}$ and Universum data $ X_{\mathfrak{U}} $;\\
			-- Decide on the total number of tasks included in the data set and assign this value to T;\\
			-- Select classification task $ S_{t}~(t=1,\ldots,T) $ in training  data set $\tilde{T}$;\\
			-- Divide Universum data $ X_{\mathfrak{U}} $ by $t$-task and get $ X_{\mathfrak{U}t}~ (t=1,\ldots,T)$;\\
			-- Choose appropriate parameters
			$c_{1}, c_{2},c_{u}$, $c_{u}^{*}$, $ \mu_{1} $, $ \mu_{2} $, and  parameter $\varepsilon \in (0,1)$.\\
		-- Select proper kernel function and kernel parameter.} \\
		{\textbf{The outputs:}}
		\begin{list}{--}{}
			\item $ u_{0},~u_{t},~v_{0}$, and $v_{t}. $
		\end{list}
		
		{\textbf{The process:}}
		
		\STATE
		Solve the two small systems of linear equations (41) and (46), and get $ \alpha,~\beta,~\alpha^{*}$, and $\beta^{*}. $
		\STATE
		Calculate  $ u_{0},~u_{t},~v_{0}$, and $v_{t}. $
		\STATE
			By utilizing the decision function (\ref{15}), assign a new point $ x $ in the $t$-th task to class $ +1 $ or $ -1 $.
	\end{algorithmic}
\end{algorithm}




\section{Numerical experiments}
This section presents the results of experiments on various single-task learning algorithms and multi-task learning algorithms. 
 The single-task learning algorithms considered consist of TBSVM \cite{shao2011improvements}, I$ \nu $-TBSVM \cite{wang2018improved} and $ \mathfrak{U}_{LS} $-TSVM \cite{xu2016least}, while the multi-task learning methods are DMTSVM \cite{xie2012multitask}, MTLS-TWSVM \cite{mei2019multi}, and our proposed methods, i.e., $\mathfrak{U}$MTSVM and LS-$\mathfrak{U}$MTSVM.
 All numerical
experiments for both linear and nonlinear models were performed in Matlab R2018b on a PC with 4 GB of RAM and Core(TM) i7 CPU @2.20 GHz under the Microsoft Windows 64-bit operating system. Moreover, to determine the classification accuracies and performance of the algorithms, we used a five-fold strategy for cross-validation. The following steps describe the cross-validation procedure.

\begin{itemize}
	\item  Partition the data sets randomly into five separate subsets of equal size.
	\item  Apply the model to four of the subsets selected as the training data.
	\item  Consider the one remaining subsets as the test data and evaluate the model on it.
	\item  Repeat the process until each of the five sets has been utilized as test data.
\end{itemize}
The accuracy is defined as the number of accurate predictions divided by the total number of forecasts. The value is then multiplied by 100 to give the percentage accuracy. We randomly selected an equal amount of data from each class for the benchmark and medical data sets, then used half of them to build the Universum data by averaging pairs of samples from different classes.
\subsection{Parameters selection}

\begin{figure}[hbt!]
	\centering
	\includegraphics[width=12cm]{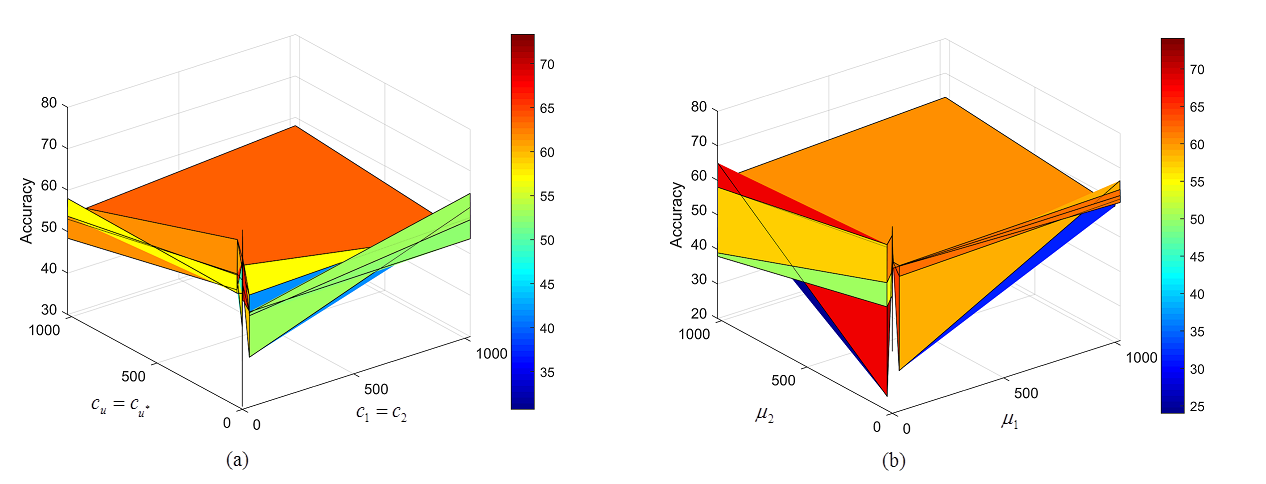}
	\caption{The effect of different values of parameter on the Caesarian data sets}
	\label{fig1}
\end{figure}

It is obvious that the performance of classification algorithms depends on the selection of proper parameters \cite{lu2018svm,xiao2021new}. Therefore, we will discuss selecting the optimal parameters of the single-task and multi-task learning methods. This subsection used five-fold cross-validation using the grid search approach to select parameters. The 3D surface plots in Figure~\ref{fig1}  demonstrate the influence of changing in the various values of parameters $c_{1}$, $c_{2}$, $c_{u}$, $c_{u^{*}}$, $\mu_{1}$ and $\mu_{2}$ on accuracy for proposed LS-$\mathfrak{U}$MTSVM method on Caesarian data set in linear state. From Figure~\ref{fig1}(a--b), it can be inferred that the accuracy obtained is quite sensitive to the selected parameters. Of course, the accuracy may be stable at some values. Therefore, the choice of parameters depends on the distribution of points in a particular data set.

The performance of the particular algorithms is determined by the parameters  $ c_{1}$, $ c_{2}$, $ c_{3}$ and $c_{4}$ in TBSVM; $ c_{1}$, $ c_{2}$, $ c_{3}$, $ c_{4}$  and $ \nu_{1},  \nu_{2} $  in I$ \nu $-TBSVM; $ c_{1}$, $ c_{2}$, $ c_{u}$, $c_{u}^{*}$ and $ \varepsilon $  in $ \mathfrak{U}_{LS} $-TSVM; $ c_{1}$, $c_{2}$, $\mu_{1}$ and $\mu_{2}$ in DMTSVM; $ c_{1},~c_{2},~\mu_{1}$ and $\mu_{2}$ in MTLS-TWSVM; $ c_{1}$, $c_{2}$, $c_{u}$, $c_{u}^{*}$, $\mu_{1}$, $\mu_{2}$  and $ \varepsilon $ in $\mathfrak{U}$MTSVM; and $ c_{1}$, $c_{2}$, $c_{u}$, $c_{u}^{*}$, $\mu_{1}$, $\mu_{2}$ and $ \varepsilon $ in LS-$\mathfrak{U}$MTSVM. Therefore, the following ranges are considered for selecting the optimal values of the parameters. In our experiments, parameters $ c_{1}$, $c_{2}, c_{3}$, $c_{4}$, $c_{u}$, $c_{u}^{*}$, $\mu_{1}$ and $\mu_{2} $ are all selected from the set $ \lbrace 2^i \mid i=-10,\dots,10\rbrace $; $  \nu_{1}$,  $\nu_{2}$ and $ \varepsilon $ are selected from the set $ \lbrace 0.1,\dots,0.9\rbrace $.

In our experiments, due to the better performance for inseparably data sets, we use the Gaussian kernel function (that is, $K(x,y)=\exp(-\gamma \|x-y\|^2)$, $\gamma > 0$).
We choose a value for the kernel parameter $ \gamma  $ from the range $ \lbrace 2^i \mid i=-10,\dots,10\rbrace$.

\subsection{Benchmark data sets}
In this subsection, we have compared multi-task learning algorithms on five benchmark data sets involving: Monk, Landmine, Isolet, Flags and Emotions. Table~\ref{tab1} shows details of data sets. 

\begin{table}[htp!]
	\small
		\centering
	\caption{The information of benchmark data sets.}\label{tab1}
	\begin{tabular}{cccc}
		\toprule
		Data set   & $\#$ Samples & $\#$ Features & $\#$ Tasks\\
		\midrule
		Monk   &432 &6 & 3 \\
		Landmine  &9674 &9 & 4 \\
		Isolet& 7797&617 & 5 \\
		Flags   & 194& 19 & 7\\
		Emotions   &593 & 72 & 6 \\
		\bottomrule
	\end{tabular}
\end{table}
The information of these data sets is as follows. 

\begin{itemize}

	\item \textbf{Monk}: In July 1991, the monks of Corsendonk Priory were confronted with the 2nd European Summer School on Machine Learning, which was hosted at their convent. After listening to various learning algorithms for more than a week, they were confused: which algorithm would be the best? Which ones should you stay away from? As a consequence of this quandary, they devised the three MONK's challenges as a basic goal to which all learning algorithms should be compared.
	\item \textbf{Landmine}: This data collection was gathered from 29 landmine regions, each of which corresponded to a separate landmine field. The data set comprises nine characteristics, and each sample is labeled with a 1 for a landmine or a 0 for a cluster, reflecting positive and negative classifications. The first 15 areas correlate too strongly with foliated regions, whereas the latter 14 belong to bare ground or uninhabited places. We adopted the following procedures for the Landmine data set to get better and more equitable experimental outcomes. This data set has more negative labels than positive ones; as a result, we started by removing some negative samples to balance things out. In our experiment, we partition the data into four tasks.
As a result, we picked four densely foliated areas as a selection of positive data. We constructed an experimental data set using four places from bare ground or desert regions as a negative data subset. To produce the data set represented by Landmine in Table~\ref{tab2}, we selected the four sites 1, 2, 3, and 4 from foliated regions and identified the four areas 16, 17, 18, and 19 from bare earth regions.

	\item\textbf{Isolet}: Isolet is a widely used data set in speech recognition that is collected as follows. One hundred fifty peoples speak 26 letters of the English alphabet twice. Therefore, 52 training samples are generated for each speaker.  Each speaker is classified into one of five categories. Consequently, we have five sets of data sets that can be considered five tasks.
	On the one hand, these five tasks are closely related because they are gathered from the same utterances. On the other hand, the five tasks differ because speakers in different groups pronounce the English letters differently.
	In this paper, we selected four pairs of similar-sounding letters, including  (O, U), (X, Y), (H, L) and (P, Q) for our experiments.
	\item \textbf{Flags}: The flags data set offers information about the flags of various countries. It contains 194 samples and 19 features. This data set is divided into seven tasks based on different colors. Each task is represented by a 19-dimensional feature vector derived from flag images of several nations. 
	Each sample may have a maximum of seven labels, because the task of recognizing each label may be seen as connected. Hence, we consider it as a multi-task learning problem. Thus, this data set contains seven tasks. In Table~\ref{tab2}, we compare the performance of the aforementioned multi-task learning methods on this data set.
	\item\textbf{Emotions}: Emotion recognition from text is one of the most difficult challenges in Natural Language Processing. The reason for this is the lack of labeled data sets and the problem's multi-class character. Humans experience a wide range of emotions, and it is difficult to gather enough information for each feeling, resulting in the issue of class imbalance. We have labeled data for emotion recognition here, and the goal is to construct an efficient model to identify the emotion.
The Emotions data collection comprises Twitter posts in English that depict six primary emotions: anger, contempt, fear, joy, sorrow, and surprise. All samples are labeled in six different ways. Each sample may contain more than one label (or emotion). Different emotion recognition tasks have similar characteristics and can be considered related tasks.  So it may be viewed as a multi-task classification issue, with each task requiring the identification of a single kind of emotion. We use 50 samples from this data set to test various multi-task learning algorithms in this experiment. The results of comparing the performance of multi-task learning algorithms on this data set are reported in Table~\ref{tab2}.
\end{itemize}
As mentioned, we compare the performance of the proposed methods with DMTSVM and MTLS-TWSVM in this subsection. Table~\ref{tab2} shows the average accuracies (``Acc''), standard deviations (``Std'') and the running time (``Time'')  on the five popular multi-task data sets. 
As is seen in Table~\ref{tab2}, the best performance is achieved by the proposed LS-$\mathfrak{U}$MTSVM followed by $ \mathfrak{U}$MTSVM. For example, on the Emotions data set, the accuracies for DMTSVM and MTLS-TWSVM  are $65.33\%$, and $66\%$, respectively. In comparison, the  $ \mathfrak{U}$MTSVM and LS-$\mathfrak{U}$MTSVM method achieved the accuracies $69.30\%$, and  $75.73\%$, which performs better than the other two multi-task learning algorithms. As a result, due to the nature of multi-task learning, it is advantageous to combine data with Universum data throughout the training phase.
Furthermore, the results seem to match our intuition that Universum data play an essential role in the performance of $ \mathfrak{U}$MTSVM and LS-$\mathfrak{U}$MTSVM. When $ \mathfrak{U}$MTSVM and LS-$\mathfrak{U}$MTSVM compare with DMTSVM and MTLS-TWSVM, we find that our proposed algorithms indeed exploit Universum data to improve the prediction accuracy and stability.
Therefore, $ \mathfrak{U}$MTSVM and LS-$\mathfrak{U}$MTSVM perform better than other  multi-task learning algorithms, i.e., DMTSVM and MTLS-TWSVM. 

In terms of learning speed, although our proposed methods are not the fastest ones due incorporating Universum data,  they offer better accuracies at an acceptable time.

\begin{table}[htp!]
	\small
	\caption{Performance comparison of nonlinear  multi-task learning methods on benchmark data sets.}\label{tab2}
	\begin{tabular}{ccccc}
		\toprule
		 & DMTSVM& $ \mathfrak{U} $MTSVM& MTLS-TWSVM & LS-$\mathfrak{U}$MTSVM\\
		Data set  & Acc ($\%$)$ \pm $Std &Acc ($\%$) $ \pm $Std& Acc ($\%$)$ \pm $Std & Acc ($\%$)$ \pm $Std \\
		& Time ($s$)&Time ($s$)& Time ($s$)& Time ($s$)\\
		\midrule
		Monk&92.76$ \pm $0.02&99.61$ \pm $0.00&94.80$ \pm $0.02&\textbf{99.80$ \pm $0.02}\\
		&27.82& 36.99&25.46&30.21\\
		Landmine&92.33$ \pm $0.01&93$ \pm $0.11 &94.05$ \pm $0.05&\textbf{94.50$ \pm $0.00}\\
		&40.69&42.21 &36.69&39.88\\
		Isolet (O, U) &99.60$ \pm $0.00&99.67$ \pm $0.01 &99$ \pm $0.02&\textbf{99.77$ \pm $0.01}\\
		&11.49&15.86&10.21&11.54\\
		Isolet (X, Y) &99.50$ \pm $0.07&98.83$ \pm $0.01 &99.60$ \pm $0.00&\textbf{100$ \pm $0.00}\\
		&11.60&15.28 &10.39&11.99\\
		Isolet (H, L) &98.16$ \pm $0.02&\textbf{100$ \pm $0.04} &99.83$ \pm $0.01&\textbf{100$ \pm $0.00}\\
		&11.55&15.37&10.27&11.89\\
		Isolet (P, Q) &96.83$ \pm $0.04&96.33$ \pm $0.05 &97.83$ \pm $0.03&\textbf{100$ \pm $0.00}\\
		&12.24&15.26 &10.32&11.88\\
		Flags &55.29$ \pm $0.25&57.48$ \pm $0.07 &57.13$ \pm $0.26&\textbf{59.57$ \pm $0.28}\\
		&3.49& 3.88&2.52&3.12\\
		Emotions&65.33$ \pm $0.19&69.30$ \pm $0.21 &66$ \pm $0.23&\textbf{75.73$ \pm $0.18}\\
		&2.25& 2.60&1.15&2.30\\
		\bottomrule
	\end{tabular}
\end{table}
\subsection{Medical data sets}
In this subsection of our experiments, we focus on comparing our proposed methods and several classifier methods, including single-task and multi-task learning methods in linear and nonlinear states. Therefore, we select four popular medical data sets to test these algorithms, including  Immunotherapy,  Ljubljana Breast Cancer, 	Breast Cancer Coimbra,  and Caesarian data sets. 
A summary of the data sets information is provided in Table~\ref{tab3}. The details of the data sets are as follows.

\begin{table}[htp!]
	\small
	\caption{The information of medical data sets.}\label{tab3}
		\centering
	\begin{tabular}{cccc}
		\toprule
		Data set   & $\#$ Samples & $\#$ Features & $\#$ Tasks\\
		\midrule
		Immunotherapy    &90 & 8& 3 \\
		Ljubljana Breast Cancer   &286 & 9& 5 \\
		Breast Cancer Coimbra &116 & 9 & 3 \\
		Caesarian   &80 & 5 & 2 \\ 
		\bottomrule
	\end{tabular}
\end{table}

\begin{itemize}
	\item \textbf{Immunotherapy}: This data collection provides information regarding wart treatment outcomes of 90 individuals utilizing Immunotherapy. The Immunotherapy data set includes 90 instances, and each instance has eight features. The features of this data include sex, age,  type, number of warts, induration diameter, area and the result of treatment.
	For this data set, we partition the data into three tasks using the variable type: task 1 (type \quo{1} = Common, 47 instances), task 2 (type \quo{2} = Plantar, 22 instances), and task 3 (type \quo{3} = Both, 21 instances). Since the kind of wart within each job are varied, this variable is also incorporated in our model.
	
	\item \textbf{Ljubljana Breast Cancer}: Nowadays, breast cancer is one of the most common malignancies in women, which has captured the attention of people all around the globe. The illness is the leading cause of mortality in women aged 40 to 50, accounting for around one-fifth of all fatalities in this age range. Every year, more than 14,000 individuals die, and the number is increasing \cite{wang2021data}. Thus, there remains a need to remove the cancer early to reduce recurrence. Since recurrence within five years of diagnosis is correlated with the chance of death, understanding and predicting recurrence susceptibility is critical. The Ljubljana breast cancer data set provides 286 data points on the return of breast cancer five years following surgical removal of a tumor. After deleting nine instances where values are missing, we are left with 277.  Each data point has one class label (for recurrence or no-recurrence events) and nine attributes, including age, menopausal status, tumor size, invasive nodes, node caps, degree of malignancy, breast (left, right), breast quadrant (left-up, left-low, right-up, right-low, central), and irradiation (yes, no).
	Here, we divide the data into five tasks using the variable tumor size: task 1 (0 $ \leq $tumor size$ \leq $19), task 2 (20 $ \leq $ tumor size $ \leq $24), task 3 (25 $ \leq $ tumor size $ \leq $29), task 4 (30 $ \leq $ tumor size $ \leq $34), and task 5 (35 $ \leq $tumor size $ \leq $54).  
	
	\item \textbf{Breast Cancer Coimbra}: The Breast Cancer Coimbra data set is the second breast cancer data set we utilize for comparison. The Gynecology Department of the Coimbra Hospital and University Center (CHUC) in Portugal collected this data set between 2009 and 2013. The Breast Cancer Coimbra data set contains 116 instances, each with nine features. This data set consists of 9 quantitative attributes and a class label attribute indicating if the clinical result is positive for existing cancer or negative (patient or healthy). Clinical characteristics were observed or assessed in 64 patients with breast cancer and 52 healthy controls. Age, BMI, insulin, glucose, HOMA, leptin, resistin, adiponectin,  and MCP-1 are all quantitative characteristics. The features are anthropometric data and measurements acquired during standard blood analysis. The qualities have the potential to be employed as a biomarker for breast cancer.
	In this experiment, we partition the data set into three tasks using the feature BMI. Based on tissue mass (muscle, fat, and bone) and height, the BMI is a simple rule of thumb to classify a person as underweight, normal weight, overweight, or obese. Underweight (less than 18.5 $kg/m^2$), normal weight (18.5 $kg/m^2$ to 24.9 $kg/m^2$), overweight (25 $kg/m^2$ to 29.9 $kg/m^2$), and obese (30 $kg/m^2$ or more) are the four major adult BMI categories.
	Hence, we consider that the first task is underweight people, the second task is normal-weight people, and the third task is overweight and obese.
	
	\item\textbf{Caesarian}: This data set, which aims to deliver via cesarean section or natural birth, provides information on 80 pregnant women who have had the most extreme delivery complications  in the medical field. The Caesarian data set includes 80 instances, and each instance has
	five features. The features of this data include age, delivery number, blood pressure, delivery time,  and heart problem. The heart problem
	feature in Caesarian data sets has two forms. We separate the data into two tasks using the variable a heart problem; task 1: The patient has
	a heart problem, and task 2: The patient does not have a heart problem.
\end{itemize}
To analyze the performance of our proposed methods, we used medical data sets and compared our proposed algorithms to five (multi-task and single-task learning) algorithms, i.e.,  TBSVM, I$ \nu $-TBSVM, $ \mathfrak{U}_{LS}$-TSVM, DMTSVM, and MTLS-TWSVM.
 We can see from the results of Tables~\ref{tab4} and~\ref{tab5} that our algorithms outperform all algorithms in linear and nonlinear states.  This occurs because proposed methods add Universum data to the model learning process to modify the classification decision boundaries. The proposed methods train all tasks simultaneously, and they can take advantage of the underlying information among all tasks and improve their performance.

\begin{table}[htp!]
	\small
	\caption{Performance comparison of linear single-task and multi-task learning methods on medical data sets.}\label{tab4}
	\scalebox{0.85}{
	\begin{tabular}{ccccc}
		\toprule
		  & Immunotherapy& 	Ljubljana Breast Cancer & 	Breast Cancer Coimbra &Caesarian \\
		Methods  & Acc ($\%$)$ \pm $Std &Acc ($\%$) $ \pm $Std& Acc ($\%$)$ \pm $Std & Acc ($\%$)$ \pm $Std \\
		& Time ($s$)&Time ($s$)& Time ($s$)& Time ($s$)\\
		\midrule
		TBSVM&77.81$ \pm $0.09 &75.11 $ \pm $0.06&72.36$ \pm $0.02&69.09$ \pm $0.03 \\
		&1.41 &1.49 &1.40&1.42 \\
		I$\nu$-TBSVM &79.97$ \pm $0.15& 73.30$ \pm $0.03&73.26$ \pm $0.15&65.01$ \pm $0.06 \\
		&1.63&3.22 &1.76& 1.58\\
		$ \mathfrak{U}_{LS} $-TSVM&78.92$ \pm $0.10&  74.72$ \pm $0.08   &79.31$ \pm $0.09&71.59$ \pm $0.11\\
		&0.04&0.04&0.45&0.04\\
		DMTSVM &   81.29$ \pm $0.06&71.14$ \pm $0.09 &75.43$ \pm $0.20&63.86$ \pm $0.12\\
		&1.49& 1.60&1.50&1.48\\
		$ \mathfrak{U} $MTSVM &84.63$ \pm $0.07&73.26$ \pm $0.08 &80.24$ \pm $0.13&71.67$ \pm $0.13\\
		&1.53&1.64 &1.55&1.48\\
		MTLS-TWSVM &86.11$ \pm $0.08&75.09$ \pm $0.19 &83.39$ \pm $0.11&76.95$ \pm $0.11\\
		&0.20&0.25 &0.19&0.16\\
		LS-$\mathfrak{U}$MTSVM &\textbf{88.11$ \pm $0.11}& \textbf{75.41$ \pm $0.00}  &\textbf{85.37$ \pm $0.12}&\textbf{78.52$ \pm $0.12}\\
		&0.25& 0.33&0.25&0.20\\
		\bottomrule
	\end{tabular}}
\end{table}
\begin{table}[htp!]
	\small
	\caption{Performance comparison of nonlinear single-task and multi-task learning methods on medical data sets.}\label{tab5}
\scalebox{0.85}{
	\begin{tabular}{ccccc}
			\toprule
		  & Immunotherapy& Ljubljana Breast Cancer & Breast Cancer Coimbra &Caesarian \\
		Methods  & Acc ($\%$)$ \pm $Std &Acc ($\%$) $ \pm $Std& Acc ($\%$)$ \pm $Std & Acc ($\%$)$ \pm $Std \\
		& Time ($s$)&Time ($s$)& Time ($s$)& Time ($s$)\\
			\midrule
		TBSVM&78.88$ \pm $0.02 &72.54$ \pm $0.05&60.39$ \pm $0.07&71.35$ \pm $0.04 \\
		&1.50 &1.49 &1.46&1.47 \\
		I$\nu$-TBSVM &78.92$ \pm $0.01& 72.90$ \pm $0.03&60.87$ \pm $0.17&71.14$ \pm $0.14 \\
		&1.43&1.59 &1.45&1.46 \\
		$ \mathfrak{U}_{LS} $-TSVM&78.92$ \pm $0.02&74.74$ \pm $0.37&63.84$ \pm $0.06&72.28$ \pm $0.02\\
		&0.06&0.11&0.07&0.07\\
		DMTSVM &80.25$ \pm $0.05&73.71$ \pm $0.09 &63.74$ \pm $0.09&72.56$ \pm $0.13\\
		&1.58& 2.14&1.67&1.56\\
		$ \mathfrak{U} $MTSVM &80.25$ \pm $0.05&75.62$ \pm $0.11 &65.71$ \pm $0.19&74.14$ \pm $0.1\\
		&1.53&2.15 &1.69&1.57\\
		MTLS-TWSVM &80.22$ \pm $0.25&75.32$ \pm $0.71 &65.33$ \pm $0.11&73.15$ \pm $0.09\\
		&0.25&0.71 &0.30&0.21\\
		LS-$\mathfrak{U}$MTSVM &\textbf{80.56$ \pm $0.05}&\textbf{77.15$ \pm $0.09} &\textbf{67.17$ \pm $0.14}&\textbf{76.74$ \pm $0.11}\\
		&0.38& 1.55&0.46&0.27\\
		\bottomrule
	\end{tabular}
}
\end{table}

\section{Conclusion}
In this paper, we introduced the twin support vector machine in the framework of multi-task learning with Universum data sets and proposed a new model called $\mathfrak{U}$MTSVM. In addition, we suggested two approaches to solving our novel model. As the first approach, we solved the $\mathfrak{U}$MTSVM by the dual problem, a quadratic programming problem. Also, we suggested the least-squares version of $\mathfrak{U}$MTSVM and called it LS-$\mathfrak{U}$MTSVM. The LS-$\mathfrak{U}$MTSVM only dealt with two systems of linear equations. Hence, comprehensive experiments on several popular multi-task data sets and medical data sets demonstrated the effectiveness of our proposed methods in terms of classification performance. The experiments confirmed that our algorithms achieved better experimental results compared to three single-task learning algorithms and two multi-task learning algorithms. 

\subsubsection*{Acknowledgments} 
H. Moosaei and M~Hlad\'{\i}k were supported by the Czech Science Foundation Grant P403-22-11117S. 
 In addition, the work of H. Moosaei was supported by the Center for Foundations of Modern Computer Science (Charles Univ.\ project UNCE/SCI/004). 
 
\section*{Conflict of interest}
The authors state that they do not have any conflicts of interest.

\bibliographystyle{spmpsci}
\bibliography{references}

\begin{thebibliography}{10}
\providecommand{\url}[1]{{#1}}
\providecommand{\urlprefix}{URL }
\expandafter\ifx\csname urlstyle\endcsname\relax
  \providecommand{\doi}[1]{DOI~\discretionary{}{}{}#1}\else
  \providecommand{\doi}{DOI~\discretionary{}{}{}\begingroup
  \urlstyle{rm}\Url}\fi

\bibitem{ando2005framework}
Ando, R.K., Zhang, T., Bartlett, P.: A framework for learning predictive
  structures from multiple tasks and unlabeled data.
\newblock Journal of Machine Learning Research \textbf{6}(11) (2005)

\bibitem{bakker2003task}
Bakker, B., Heskes, T.: Task clustering and gating for bayesian multitask
  learning.
\newblock Journal of Machine Learning Research \textbf{4} (2003)

\bibitem{bi2008improved}
Bi, J., Xiong, T., Yu, S., Dundar, M., Rao, R.B.: An improved multi-task
  learning approach with applications in medical diagnosis.
\newblock In: Joint European Conference on Machine Learning and Knowledge
  Discovery in Databases, pp. 117--132. Springer (2008)

\bibitem{birlutiu2010multi}
Birlutiu, A., Groot, P., Heskes, T.: Multi-task preference learning with an
  application to hearing aid personalization.
\newblock Neurocomputing \textbf{73}(7-9), 1177--1185 (2010)

\bibitem{burges1998tutorial}
Burges, C.J.C.: A tutorial on support vector machines for pattern recognition.
\newblock Data Mining and Knowledge Discovery \textbf{2}(2), 121--167 (1998)

\bibitem{chapelle2007analysis}
Chapelle, O., Agarwal, A., Sinz, F., Sch{\"o}lkopf, B.: An analysis of
  inference with the universum.
\newblock Advances in Neural Information Processing Systems \textbf{20},
  1369--1376 (2007)

\bibitem{chapelle2010multi}
Chapelle, O., Shivaswamy, P., Vadrevu, S., Weinberger, K., Zhang, Y., Tseng,
  B.: Multi-task learning for boosting with application to web search ranking.
\newblock In: Proceedings of the 16th ACM SIGKDD international conference on
  Knowledge discovery and data mining, pp. 1189--1198 (2010)

\bibitem{cheng2015multi}
Cheng, X., Li, N., Zhou, T., Wu, Z., Zhou, L.: Multi-task object tracking with
  feature selection.
\newblock IEICE Transactions on Fundamentals of Electronics, Communications and
  Computer Sciences \textbf{98}(6), 1351--1354 (2015)

\bibitem{evgeniou2004regularized}
Evgeniou, T., Pontil, M.: Regularized multi--task learning.
\newblock In: Proceedings of the tenth ACM SIGKDD international conference on
  Knowledge discovery and data mining, pp. 109--117 (2004)

\bibitem{khemchandani2007twin}
Jayadeva, Khemchandani, R., Chandra, S.: Twin support vector machines for
  pattern classification.
\newblock IEEE Transactions on Pattern Analysis and Machine Intelligence
  \textbf{29}(5), 905--910 (2007)

\bibitem{ji2013multitask}
Ji, Y., Sun, S.: Multitask multiclass support vector machines: model and
  experiments.
\newblock Pattern Recognition \textbf{46}(3), 914--924 (2013)

\bibitem{kumar2009least}
Kumar, M.A., Gopal, M.: Least squares twin support vector machines for pattern
  classification.
\newblock Expert Systems with Applications \textbf{36}(4), 7535--7543 (2009)

\bibitem{lu2018svm}
Lu, L., Lin, Q., Pei, H., Zhong, P.: The {aLS}-{SVM} based multi-task learning
  classifiers.
\newblock Applied Intelligence \textbf{48}(8), 2393--2407 (2018)

\bibitem{mangasarian2005multisurface}
Mangasarian, O.L., Wild, E.W.: Multisurface proximal support vector machine
  classification via generalized eigenvalues.
\newblock IEEE Transactions on Pattern Analysis and Machine Intelligence
  \textbf{28}(1), 69--74 (2006)

\bibitem{mei2019multi}
Mei, B., Xu, Y.: Multi-task least squares twin support vector machine for
  classification.
\newblock Neurocomputing \textbf{338}, 26--33 (2019)

\bibitem{moosaei2021DC}
Moosaei, H., Bazikar, F., Ketabchi, S., Hlad\'{\i}k, M.: Universum
  parametric-margin $\nu$-support vector machine for classification using the
  difference of convex functions algorithm.
\newblock Appl. Intell. \textbf{52}(3), 2634--2654 (2022)

\bibitem{moosaei2021sparse}
Moosaei, H., Mousavi, A., Hlad{\'\i}k, M., Gao, Z.: Sparse universum quadratic
  surface support vector machine models for binary classification.
\newblock arXiv preprint arXiv:2104.01331  (2021)

\bibitem{qi2012twin}
Qi, Z., Tian, Y., Shi, Y.: Twin support vector machine with universum data.
\newblock Neural Networks \textbf{36}, 112--119 (2012)

\bibitem{ren2016multicell}
Ren, Y., Xu, B., Zhu, P., Lu, M., Jiang, D.: A multicell visual tracking
  algorithm using multi-task particle swarm optimization for low-contrast image
  sequences.
\newblock Applied Intelligence \textbf{45}(4), 1129--1147 (2016)

\bibitem{richhariya2018improved}
Richhariya, B., Sharma, A., Tanveer, M.: Improved universum twin support vector
  machine.
\newblock In: S.~Sundaram (ed.) 2018 IEEE Symposium Series on Computational
  Intelligence (IEEE SSCI 2018), pp. 2045--2052. IEEE (2018)

\bibitem{richhariya2018eeg}
Richhariya, B., Tanveer, M.: {EEG} signal classification using universum
  support vector machine.
\newblock Expert Systems with Applications \textbf{106}, 169--182 (2018)

\bibitem{shao2011improvements}
Shao, Y.H., Zhang, C.H., Wang, X.B., Deng, N.Y.: Improvements on twin support
  vector machines.
\newblock IEEE Transactions on Neural Networks \textbf{22}(6), 962--968 (2011)

\bibitem{shiao2012implementation}
Shiao, H.T., Cherkassky, V.: Implementation and comparison of {SVM}-based
  multi-task learning methods.
\newblock In: The 2012 International Joint Conference on Neural Networks
  (IJCNN), pp. 1--7. IEEE (2012)

\bibitem{su2017multi}
Su, C., Yang, F., Zhang, S., Tian, Q., Davis, L.S., Gao, W.: Multi-task
  learning with low rank attribute embedding for multi-camera person
  re-identification.
\newblock IEEE Transactions on Pattern Analysis and Machine Intelligence
  \textbf{40}(5), 1167--1181 (2017)

\bibitem{vapnik2006estimation}
Vapnik, V.: Estimation of Dependences Based on Empirical Data.
\newblock Springer, New York (2006)

\bibitem{vapnik200624}
Vapnik, V.: Transductive inference and semi-supervised learning.
\newblock In: O.~Chapelle, B.~Sch\"{o}lkopf, A.~Zien (eds.) Semi-Supervised
  Learning, pp. 453--472. MIT Press (2006)

\bibitem{wang2018improved}
Wang, H., Zhou, Z., Xu, Y.: An improved $\nu$-twin bounded support vector
  machine.
\newblock Applied Intelligence \textbf{48}(4), 1041--1053 (2018)

\bibitem{wang2021data}
Wang, L.: Data science for characterizing breast cancer.
\newblock In: 2021 3rd International Conference on Intelligent Medicine and
  Image Processing, pp. 122--126 (2021)

\bibitem{weston2006inference}
Weston, J., Collobert, R., Sinz, F., Bottou, L., Vapnik, V.: Inference with the
  universum.
\newblock In: Proceedings of the 23rd International Conference on Machine
  Learning, pp. 1009--1016 (2006)

\bibitem{xiao2021new}
Xiao, Y., Wen, J., Liu, B.: A new multi-task learning method with universum
  data.
\newblock Applied Intelligence \textbf{51}(6), 3421--3434 (2021)

\bibitem{xie2012multitask}
Xie, X., Sun, S.: Multitask twin support vector machines.
\newblock In: International Conference on Neural Information Processing, pp.
  341--348. Springer (2012)

\bibitem{xie2015multitask}
Xie, X., Sun, S.: Multitask centroid twin support vector machines.
\newblock Neurocomputing \textbf{149}, 1085--1091 (2015)

\bibitem{xu2016least}
Xu, Y., Chen, M., Li, G.: Least squares twin support vector machine with
  universum data for classification.
\newblock International Journal of Systems Science \textbf{47}(15), 3637--3645
  (2016)

\bibitem{xu2016nu}
Xu, Y., Chen, M., Yang, Z., Li, G.: $\nu$-twin support vector machine with
  universum data for classification.
\newblock Applied Intelligence \textbf{44}(4), 956--968 (2016)

\bibitem{xue2016multi}
Xue, Y., Beauseroy, P.: Multi-task learning for one-class {SVM} with additional
  new features.
\newblock In: 2016 23rd International Conference on Pattern Recognition (ICPR),
  pp. 1571--1576. IEEE (2016)

\bibitem{yang2010multi}
Yang, H., King, I., Lyu, M.R.: Multi-task learning for one-class
  classification.
\newblock In: The 2010 International Joint Conference on Neural Networks
  (IJCNN), pp. 1--8. IEEE (2010)

\bibitem{zhang2008semi}
Zhang, D., Wang, J., Wang, F., Zhang, C.: Semi-supervised classification with
  universum.
\newblock In: Proceedings of the 2008 SIAM International Conference on Data
  Mining, pp. 323--333. SIAM (2008)

\bibitem{zhou2018position}
Zhou, D., Miao, L., He, Y.: Position-aware deep multi-task learning for
  drug--drug interaction extraction.
\newblock Artificial Intelligence in Medicine \textbf{87}, 1--8 (2018)

\end{thebibliography}
\end{document}